\documentclass{article}

\PassOptionsToPackage{numbers, compress}{natbib}

\usepackage[preprint]{neurips}
\usepackage{booktabs}       
\usepackage{multirow}
\usepackage{tabularx}
\usepackage{graphicx}       
\usepackage{colortbl}       
\usepackage{setspace}
\usepackage{hyperref}
\hypersetup{
    colorlinks=true,
    urlcolor=red,
}

\usepackage{graphicx}
\usepackage{subcaption}
\usepackage{float}
\usepackage[justification=justified]{caption}	
\usepackage{lscape}                                         

\usepackage{dsfont}

\usepackage[lined,ruled,linesnumbered]{algorithm2e}

\usepackage{booktabs}                   
\usepackage{multirow}

\usepackage{paralist}
\usepackage{enumitem}

\usepackage{bm}                          
\usepackage{epsfig}                      
\usepackage{graphicx}                  
\usepackage{mathtools}
\usepackage{amssymb,amsmath} 
\usepackage{wrapfig}

\usepackage{units}
\usepackage{color}

\usepackage{comment}

\usepackage{url}  

\usepackage[capitalize]{cleveref}
\crefname{section}{Sec.}{Secs.}
\Crefname{section}{Section}{Sections}
\Crefname{table}{Table}{Tables}
\crefname{table}{Tab.}{Tabs.}

\usepackage{xspace}
\usepackage[table]{xcolor}
\usepackage{setspace}

\def\eg{e.g.,~}               
\def\ie{i.e.,~}               

\DeclareMathOperator*{\argmax}{\arg\!\max}

\newlength\paramarginsize
\newlength\figmarginsize
\newlength\secmarginsize
\newlength\figcapmarginsize
\newlength\tabcapmarginsize

\newcommand{\paramargin}{\vspace{\paramarginsize}}

\newcommand{\figcapmargin}{\vspace{\figcapmarginsize}}

\newcommand{\blue}{\textcolor{blue}}

\newcommand{\orange}{\textcolor{orange}}

\newcommand{\method}{PCEvE}

\newcommand{\mpage}[2]
{
\begin{minipage}{#1\linewidth}\centering
#2
\end{minipage}
}

\newcommand{\topic}[1]
{
\paramargin\noindent \textbf{#1}
}

\newcommand{\figcaption}[2]
{
\caption{
\textbf{#1.}  
#2            
}
}

\newcommand{\secref}[1]{Section~\ref{sec:#1}}
\newcommand{\figref}[1]{Figure~\ref{fig:#1}} 
\newcommand{\tabref}[1]{Table~\ref{tab:#1}}
\newcommand{\eqnref}[1]{\eqref{eq:#1}}

\long\def\ignorethis#1{}

\newcommand{\tb}[1]{\textbf{#1}}

\def\xi{\mathbf{x}_i}

\graphicspath{{figure}, {images}, {example}}

\title{PCEvE: Part Contribution Evaluation Based Model Explanation
for Human Figure Drawing Assessment and Beyond}

\author{
Jongseo Lee\thanks{Equally contributed first authors.} 
\And 
Geo Ahn\footnotemark[1] 
\And 
Seong Tae Kim\footnotemark[2] \\
\And
Jinwoo Choi\thanks{Corresponding author.}\\
\And
Kyung Hee University, Republic of Korea\\
{\tt\small \{jong980812, ahngeo11, st.kim, jinwoochoi\}@khu.ac.kr}
}

\begin{document}
\maketitle

\begin{abstract}
For automatic human figure drawing (HFD) assessment tasks, such as diagnosing autism spectrum disorder (ASD) using HFD images, the clarity and explainability of a model decision are crucial.
%
%
Existing pixel-level attribution-based explainable AI (XAI) approaches demand considerable effort from users to interpret the semantic information of a region in an image, which can be often time-consuming and impractical.
To overcome this challenge, we propose a part contribution evaluation based model explanation (\method{}) framework.
On top of the part detection, we measure the Shapley Value of each individual part to evaluate the contribution to a model decision.
Unlike existing attribution-based XAI approaches, the \method{} provides a straightforward explanation of a model decision, \ie a part contribution histogram.
Furthermore, the \method{} expands the scope of explanations beyond the conventional sample-level to include class-level and task-level insights, offering a richer, more comprehensive understanding of model behavior.
We rigorously validate the \method{} via extensive experiments on multiple HFD assessment datasets.
Also, we sanity-check the proposed method with a set of controlled experiments. 
Additionally, we demonstrate the versatility and applicability of our method to other domains by applying it to a photo-realistic dataset, the Stanford Cars.
\end{abstract}
\section{Introduction}
\label{sec:intro}


\begin{figure*}[t]
\centering
\includegraphics[width=1.0\linewidth]{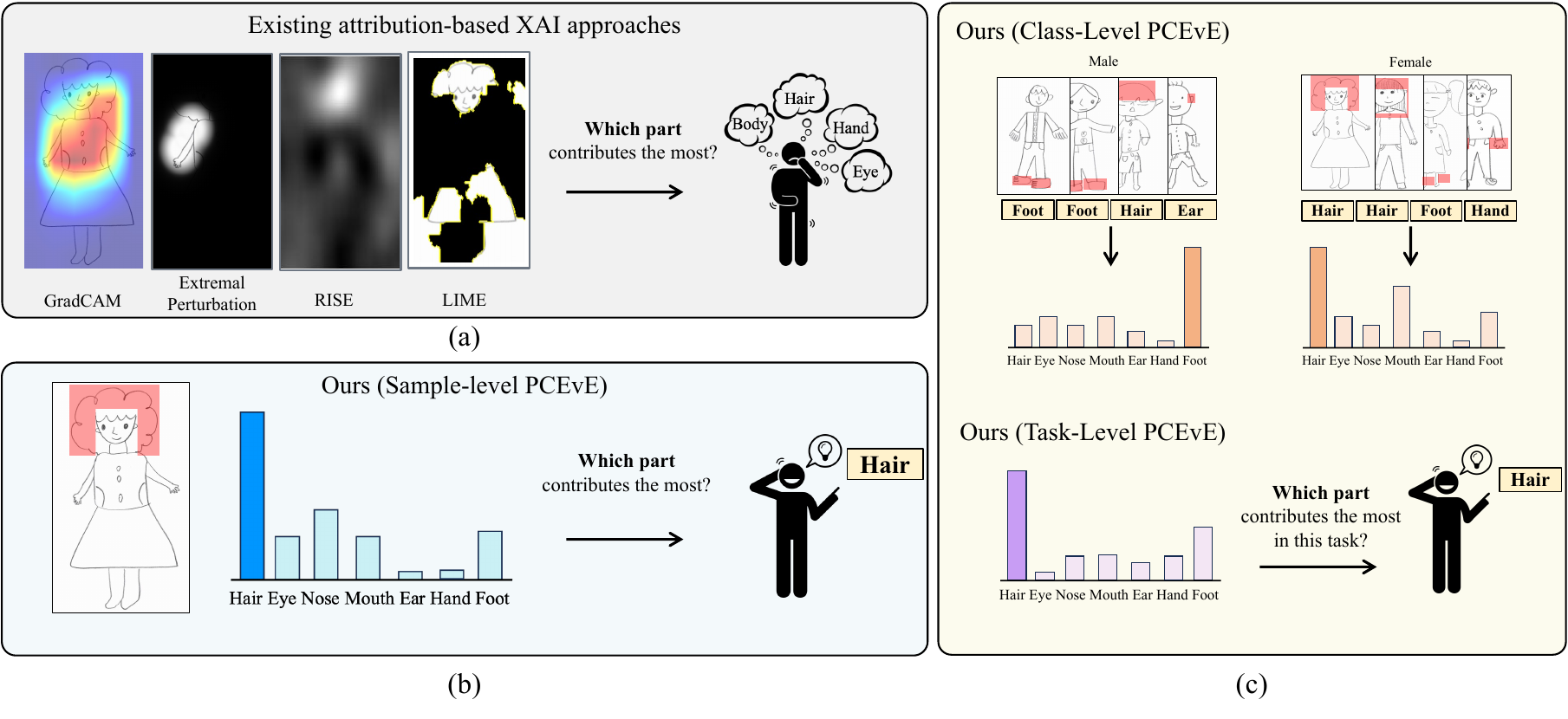}
\figcapmargin

\captionof{figure}{
\textbf{Why do we need part-based model explanations for HFD assessment?}
We show a motivating example from a drawer gender classification task to highlight the contrast between the existing XAI approaches and the proposed approach.
(a) The existing attribution-based XAI approaches visualize pixel-level attribution maps. 
However, these pixel-level attributions require users to infer which particular part is crucial in recognizing the image as `Female' for a model.
This step demands a level of interpretation that might not be immediately intuitive.
(b) In contrast, our part contribution evaluation based model explanation (\method{}) furnishes users with more direct and interpretable insights into model decisions.
The \method{} provides a part contribution histogram that eliminates the need for inference on which parts are crucial.
(c) Furthermore, the \method{} extends to provide more abstract-level insights, including class-level and task-level explanations.
With sample-level, class-level, and task-level model explanations, we can understand model decision processes across various dimensions.
}
%
\label{fig:teaser}
\end{figure*}
With recent advances in computer vision and deep learning, human figure drawing (HFD) assessment using a deep learning model has shown great progress~\cite{rakhmanov2020experimentation}.
Despite the great advancement, we still do not understand the decision-making processes underlying these models.
The transparency of a method is crucial in medical applications such as ASD diagnosis since the clear and reliable rationale behind diagnosis should be provided to subjects and their families~\cite{yu2022alzheimer}
%

In this work, we propose a novel model explanation framework for HFD assessment, referred to as Part Contribution Evaluation based model Explanation (\method{}).
Our method is designed to provide explanations based on the contributions of human body parts on an input human figure drawing image. 
With the \method{}, we can identify which part is more important than other parts in a model decision on an input human figure drawing image. 
By quantifying the contributions of parts, the \method{} provides reliable clues for a model decision, which can be integrated into the class and task level.
Part-based explanations are crucial for HFD assessment for several compelling reasons.
Parts or components serve as units of human visual perception~\cite{biederman1987recognition}.
Therefore, many explainable artificial intelligence (XAI) approaches for fine-grained recognition tasks focus on parts to explain a model decision on discriminating subtle differences among fine-grained categories
~\cite{hesse2023funnybirds}.
The HFD assessment is also a fine-grained recognition task as a model needs to distinguish subtle differences among classes.
Therefore, part-based explanations are natural in the HFD assessment.
When human experts assess HFD, the features of individual body parts are one of the key considerations~\cite{goodenough1926measurement}.
The human experts focus on the omission/inclusion or exaggeration of specific parts, or their proportions.
Therefore, part-based explanations in HFD assessment align naturally with the cognitive mechanisms of human visual perception and the established methodologies of human expert analysis.

The existing XAI approaches can provide part-related model explanations as shown in \figref{teaser} (a).
Here, we show example explanations from well-established feature attribution approaches \ie GradCAM~\cite{selvaraju2017gradcam}, External Perturbation~\cite{fong2019perturbation}, RISE~\cite{ribeiro2016lime}, and LIME~\cite{petsiuk2018rise} on a drawer gender classification task.
Each method visualizes the focus of the model by attributing the importance of each pixel for the drawer gender classification task
However, these pixel-level attributions approaches show critical limitations.
%
Sometimes the pixel-level attribution approaches fail to provide convincing explanations or are marginally effective in human-centric benchmarks~\cite{colin2022cannot}.
Furthermore, pixel-level attribution approaches force a human to interpret the semantic information of a region in an image, leading to additional costs and time for users~\cite{wu2020towards}.
In \figref{teaser} (a), the pixel-level attribution approaches do not identify \emph{which specific part} is crucial (\eg `Body' vs. `Hair' vs. `Hand' vs. `Eye') in the decision and therefore, it requires a human interpretation.  
Such a requirement underscores the need for more intuitive and part-specific explanation methodologies that bridge the gap between raw pixel-level data and the semantic understanding needed for insightful model interpretation.

To this end, we propose a novel framework, \method{} designed to enhance model explanations by shifting from pixel-level attributions to a more semantically rich understanding of parts.
The \method{} adopts the concept of the Shapley Value~\cite{shapley1953value}, a principle from cooperative game theory known for equitable contribution distribution among participants.
We apply the Shapley Value at the part-level to assess the impact of individual parts on model decisions.
On top of an off-the-shelf part detector to detect pre-defined parts, we feed all possible part combinations of an input image into the model we want to explain.
Then we measure the Shapely Value of each part of the input image.
As shown in \figref{teaser} (b), the \method{} provides more straightforward explanations of a model decision compared to the attribution-based XAI methods, \eg By the \method{}, `Hair' turns out to be the most contributing part for the model to predict the input image as `Female' class.
Our approach not only simplifies the interpretation of model decision, but also elevates the explanations to a more abstract level by aggregating part contributions across an entire class or task.
As shown in \figref{teaser} (c), the \method{} can explain that `Foot' is significant for recognizing `Male' figures, whereas `Hair' is more indicative of `Female' figures.
Also, the \method{} can explain the model focuses on `Hair' when distinguishing `Male' and `Female'.

To validate the effectiveness of our approach, we conduct extensive experiments across various HFD assessment datasets with diverse models. 
Through the experiments, we show that the \method{} pinpoints the model's focus on specific parts, providing insights that are more straightforward compared to those offered by traditional attribution-based XAI approaches.
We rigorously examine the validity of the explanation provided by the proposed method via a set of controlled experiments.
Moreover, we move beyond the HFD assessment and apply the \method{} on a photo-realistic fine-grained visual categorization dataset, Stanford Cars~\cite{krause2013car}.
Through the experiments, we showcase that our approach is not limited to HFD but provides a reasonable model explanation on a photo-realistic dataset.

In this work, we make the following key contributions that advance the field of model explanation.
\begin{itemize}
\item We propose a novel model explanation framework, referred to as \method{}, designed to explain a model decision by evaluating the contributions of individual parts of an input image.
To the best of our knowledge, the \method{} is the first approach to explain a model decision based on part contribution statistics in the HFD assessment.

\item The proposed \method{} dissects model explanations across three dimensions: individual samples, classes, and tasks.
To the best of our knowledge, we are the first to explain a model at class and task level.

\item Through extensive experiments, we demonstrate the applicability of \method{} across different human figure drawing datasets. 
Furthermore, we validate the sanity of the proposed explanation framework by a set of controlled experiments.

\item We validate the proposed method to a photo-realistic fine-grained visual categorization dataset as well.
The proposed method reasonably explains a model decision on such photo-realistic as well as the human figure drawing assessment task.

\end{itemize}

\section{Related Work}
\label{sec:related}

\subsection{Attribution-based Model Explanation}
Attribution-based model explanation approaches measure the contribution of each pixel in various ways and visualize the contribution by overlaying a heatmap on the input image as shown in \figref{teaser}(a). 
There are two main categories of attribution-based approaches: gradient-based and perturbation-based approaches.
Gradient-based attribution approaches~\cite{selvaraju2017gradcam,zheng2022highres} 
leverage the gradient of the model's output of the target class with respect to a particular feature. 
A higher gradient value suggests that a small change in the feature would lead to a significant change in the output, highlighting the contribution to the model decision. 
Perturbation-based approaches~\cite{ribeiro2016lime,fong2019perturbation} perturb one or more input features (e.g., pixels) and explain how the model prediction changes in the target class. 
By perturbing input pixels, these methods can identify regions in the input that highly contribute to the model predictions.
%
%

Attribution-based approaches can provide pixel-level explanations of a model. 
However, pixel-level attribution approaches force a human to interpret the semantic information of a region in an image, leading to additional costs~\cite{wu2020towards}.
Interpreting a saliency map is often nontrivial and may even introduce human confirmation bias through qualitative evaluation~\cite{dabkowski2017real}. 
In contrast, our work addresses the limitations of attribution-based approaches by taking a part-based model explanation approach, which offers a more intuitive framework for interpreting model decisions.

\subsection{Shapley Value}
The Shapley Value~\cite{shapley1953value} is a concept borrowed from the cooperative game theory.
In the game theory, the Shapley Value is a metric to fairly distribute payoffs among players based on their contribution to the total gain.
In machine learning, the Shapley Value offers a principled approach to quantify the importance of individual input features.
In many recent works on XAI~\cite{zheng2022shap,ahn2023line}
, the Shapely Value is used to measure the marginal contribution of each feature or neuron by considering all possible combinations of features, offering a comprehensive view of how each input impacts the model decision. 
While prior works focus on explaining a model by observing the \emph{feature or neuron}, our work focuses on explaining a model behavior by observing the \emph{part contribution}.
Moreover, our \method{} can provide a model explanation at various levels: sample, class, and task while the prior works mostly focus on sample-level explanations.
%

\subsection{Human Figure Drawing Assessment} 
There have been extensive works on art psychotherapy using a human figure drawing (HFD) assessment aimed at estimating the mental developmental status of a child.
The most popular HFD assessment tasks are Draw-a-Person (DAP) test~\cite{goodenough1926measurement} and House-Tree-Person (HTP) test~\cite{buck1948htp}.
The DAP test measures a child's intellectual developmental status through their drawing of a person. 
In the DAP test, the criteria include the presence of parts such as eyes and legs, the appropriateness of the position and proportion of each part, and other details reflecting the overall appearance~\cite{goodenough1926measurement}.
The HTP test evaluates aspects of a participant's personality, emotions, and attitudes using the drawing of a house, a tree, and a person. 
The criteria include the shape, position, size, and shading of each component, as well as the overall mood, harmony, structure of the entire picture, and line pressure, among others~\cite{buck1948htp}.
The \emph{details of each part} are important in the two popular HFD assessment tasks.

Recent works~\cite{pan2022automated,rakhmanov2020experimentation}
have shown that deep neural networks can learn the HFD assessment tasks.
There are only a few works on the deep learning based HFD assessment model explanation such as applying the CAM \cite{zhou2016cam} to the HFD assessment models~\cite{kim2024hfdxai}, and detecting objects~\cite{kim2023alphadapr} for the Drawing-A-Person-in-the-Rain assessment task~\cite{verinis1974dapr}.
However, the prior works do not provide a straightforward model explanation that a human can easily understand.
In contrast, we propose the \method{} to explain a model decision for HFD assessment tasks, which provides a straightforward model explanation: part contribution histogram.

\subsection{Concept-based and Part-based Model Explanation}
Concept-based and part-based model explanation approaches are popular in explaining fine-grained visual categorization (FGVC) model behaviors.
Concept-based approaches~\cite{fel2023craft,andres2024eclad}
extract high-level concepts learned by models. 
However, concept-based approaches heavily rely on the good representations learned by the model, making it challenging to extract reasonable concepts in data-scarce domains such as sketch-based HFD assessment tasks.
Part-based XAI methods~\cite{huang2016part,hesse2023funnybirds}
try to explain models by clarifying the semantic importance of different parts in fine-grained visual categorization tasks. 
However, it is nontrivial to apply the prior part-based XAI methods to the HFD assessment tasks. 
Since the HFD datasets often have limited size, the extracted features by a model are not quite reliable for the existing part-based XAI methods. 
To address the challenge, we define semantic parts by using part annotations/detections and evaluate part contributions using Shapley Value~\cite{shapley1953value} instead of conventional optimization.

\begin{figure}[t]
    \centering
    \includegraphics[width=0.85\linewidth]{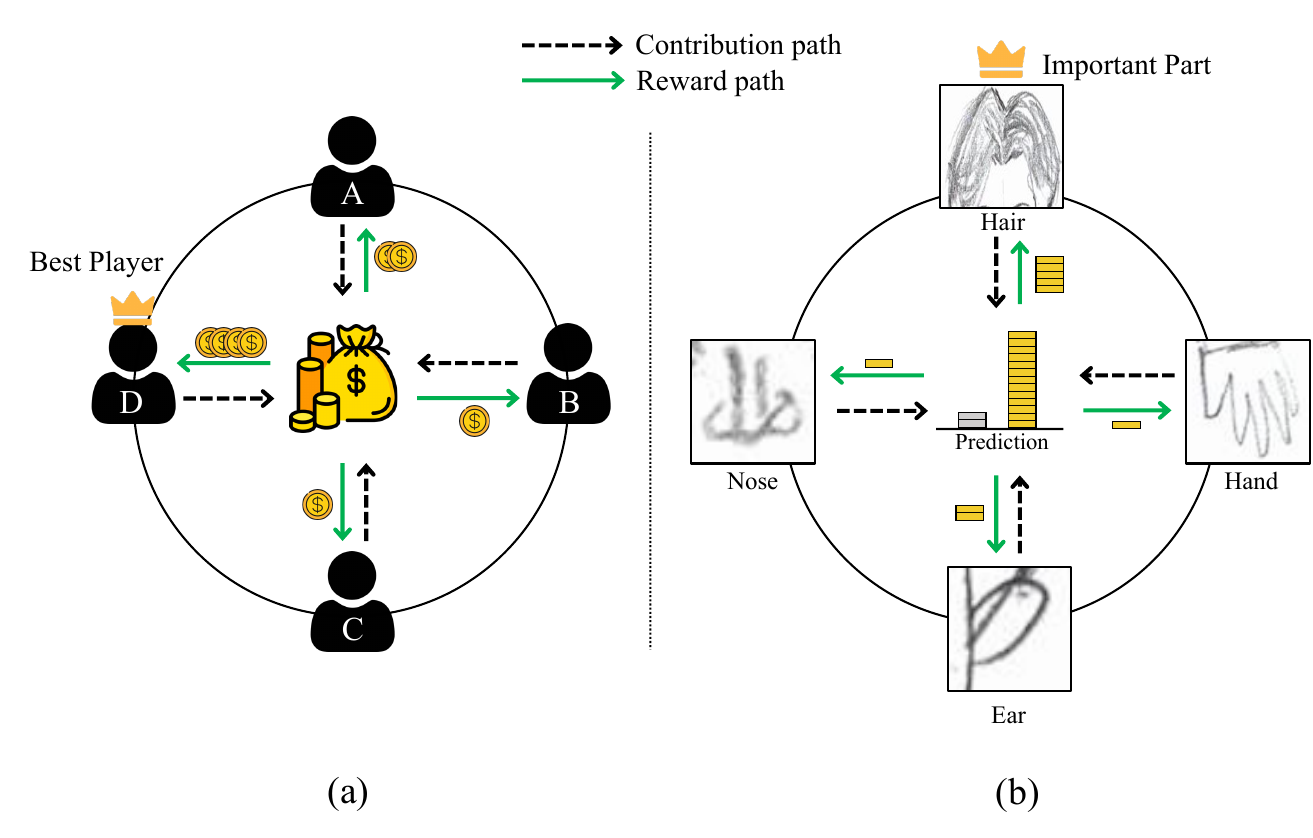}

    \figcaption{Analogy between PCEvE and game theory}{(a) Game theory: by the coalition, players obtain a certain amount of overall gain. 
    The Shapley Value ensures a fair distribution of payoff among players. 
    In this example, player D contributes the most and receives the highest reward. 
    (b) PCEvE: In this example, each body part corresponds to an individual player in (a). 
    Each body part contributes to a model prediction, and the Shapley Value allows us to quantify the contribution of each body part.}
    \vspace{-1em}
    \label{fig:motivation}
\end{figure}

\section{Part Contribution Evaluation Based Model Explanation}
\label{sec:method}

\begin{figure*}[t]
\centering
\includegraphics[width=\linewidth]{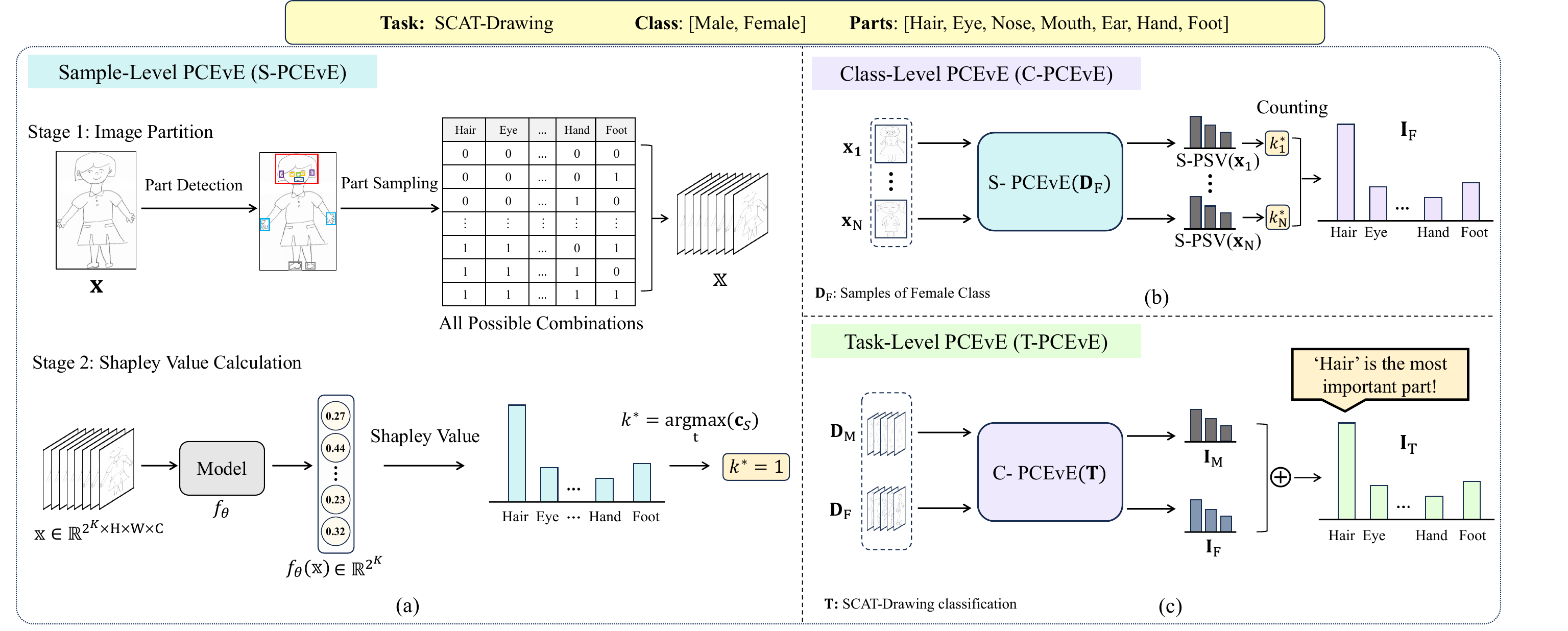}
\figcapmargin


\figcaption{Overview of PCEvE}{
The PCEvE explains a model decision by providing part contribution statistics at a sample/class/task level.
(a) Sample-level PCEvE: given an input image with $K$ parts, we generate $2^K$ images by masking each part to obtain a set of all possible part combinations: $\mathbb{X}$.
To evaluate the contribution of each part, the S-PCEvE aggregates the logit vectors of every image in $\mathbb{X}$ predicted by a model, resulting in a sample-level part contribution histogram.
(b) Given a target class, \eg{`Female'}, the C-PCEvE counts the most significant part for every image belonging to the class resulting in a class-level part contribution histogram.
(c) The T-PCEvE accumulates the class-level part contribution histograms of all classes in the task.
The T-PCEvE gives the task-level part contribution histogram, providing a model explanation at a task level.
}
\vspace{-1em}
\label{fig:method}
\end{figure*}

We introduce the Part Contribution Evaluation based model Explanation (\method{}) framework. 
As shown in \figref{method}, the \method{} provides part contribution statistics of a model at a sample/class/task level, \eg at the task level, `Hair' of the drawing contributes most to a gender classification model decision.
To evaluate part contribution, the \method{} measures the Shapley Value~\cite{shapley1953value} of each part.
Since we can measure the Shapley Value for every model, the \method{} is model-agnostic for the classification task. 
In this section, we provide a detailed description of the Shapley Value in \secref{shap}. 
Then we show the overall \method{} framework in \secref{partshap}.
We illustrate the sample-level \method{} in \secref{sample}.
Finally, we describe how we extend the sample-level \method{} to class-level and task-level \method{} in \secref{group}.
%
%

\subsection{Preliminaries: Shapley Value}
\label{sec:shap}
We provide preliminary descriptions on the Shapley Value~\cite{shapley1953value}. 
The Shapley Value is a metric to evaluate the contribution of a player in a coalitional game.
Using the Shapley Value, we can allocate a fair reward to each player by considering all the contributions in a game. 
The Shapley value is a robust metric for evaluating fair contributions among players.

Given a set, $\mathbb{N}$, of $n$ players, a value function $f:2^{\mathbb{N}}\rightarrow \mathbb{R}$ assigns each a subset $\mathbb{S}\subseteq\mathbb{N}$ a real number, where the $2^{\mathbb{N}}$ denotes the power set of $\mathbb{N}$. 
The $f(\mathbb{S})$ evaluates the summation of rewards obtained by the members of $\mathbb{S}$ in a coalitional game.
Then the Shapley Value of an $i$-th player, $\psi_i(f)$, measures the average marginal contribution that the player makes across all possible coalitions $\mathbb{S}$ containing the player $i$.
\begin{equation}
\label{eq:shap}
    {\psi _{i}(f)=\sum _{\mathbb{S}\subseteq \mathbb{N}\setminus \{i\}}{\frac {|\mathbb{S}|!\;(n-|\mathbb{S}|-1)!}{n!}}(f(\mathbb{S}\cup \{i\})-f(\mathbb{S}))}.
\end{equation}
With the Shapley Value, we can obtain a fair distribution of the total surplus from all the players by considering the pure contribution of each player in a coalitional game. 
There are four axiomatic characterizations to prove the fairness of the Shapley Value. 

\topic{Dummy Player.}
If a player has no contribution to the game, the Shapley Value for that player should be zero. 
This means the concept that players who do not contribute should not receive any payoff.

\topic{Efficiency.}
%
The sum of payoff distributed to all players must equal the total value generated by the coalitional game \ie{$f(\mathbb{N})=\sum_{i\in\mathbb{N}} \psi_i(f)$}. This ensures that the collective rewards allocated based on individual contributions match the overall value created, highlighting that no value is wasted.

\topic{Symmetry.}
If two players have the same contributions, their Shapley Values should be the same. 
This reflects the idea that players with equivalent roles should receive equivalent rewards. 

\topic{Linearity.}
If there are two coalitional games with value functions $f$ and $g$, the Shapley Value for a player in a combined game $\psi_i(a\cdot f+b\cdot g)$ is a linear combination of the two Shapley Value from individual Shapley values $\psi_i(f)$ and $\psi_i(g)$. 
\ie $\psi_i(a\cdot f+b\cdot g) = a\psi_i(f)+b\psi_i(g)$.

The Shapley Value \eqnref{shap}, satisfies all four axioms.
For these reasons, many researchers use the Shapley Value in the field of eXplainable Artificial Intelligence (XAI) to measure the contributions of various elements in a black box model~\cite{ghorbani2020neuron,zheng2022shap}. 
%
Since the Shapley Value is a fair and model-agnostic metric, we use it in \method{} to measure the contribution of each part to a model decision.



\subsection{\method{} Framework Overview}
\label{sec:partshap}
In our Part Contribution Evaluation based model Explanation (\method{}) framework, we measure the Shapley Value~\cite{shapley1953value} of each part to evaluate the contribution to a model decision on the target task.
Since the Shapley Value fairly measures contributions from all players in a coalition game, we use it to \emph{fairly} measure how much each part contributes to a model decision.
In \figref{motivation}, we provide an analogy between the \method{} and the game theory.
In the example shown in \figref{motivation} (b), we treat each part, \ie `Hair', `Nose', `Ear', and `Hand', of an input image as a player in a coalitional game, \ie a model decision.
In the example, while each body part contributes to a model decision, `Hair' contributes the most.

Let us consider a classification task with a dataset $\mathbb{D}=\{(\mathbf{x_i},y_i)\}_{i=1}^N$, where $\mathbf{x}_i$ is the $i$-th image and $y_i$ is the corresponding label, and $N$ denotes the number of samples.
As shown in \figref{method}, given an input image $\mathbf{x}_i$, we run an off-the-shelf part detector to obtain $K$ part pseudo-labels.
The part pseudo-labels are the bounding box coordinates of each part.
Then we generate $2^K$ images by masking each part to obtain a set of all possible part combinations $\mathbb{X}_i$.
In the sample-level \method{} (S-\method{}), to evaluate the contribution of each part, we aggregate the logit vectors of every image in $\mathbb{X}_i$, predicted by a model of our interest.
On top of the S-\method{}, we can obtain a group-level explanation of a model.
Given a target class, \eg `Female', the class-level \method{} (C-\method{}) counts the most significant part for every image belonging to the class, resulting in a class-level part contribution histogram.
The task-level \method{} (T-\method{}) accumulates the class-level part contribution histograms of all classes in the task to obtain the task-level part contribution.

\subsection{Sample-Level \method{}}
\label{sec:sample}
%
Given a set of all possible part combinations $\mathbb{X}_i$ of an input image $\mathbf{x}_i$, the goal of S-\method{} is to quantify the contribution of each part of the input image to the decision of a model of interest, $f$. 
Here, a model of interest $f_{\theta}$, with parameters $\theta$, serves as a value function as described in \secref{shap}. 
To explain the decision of a model $f_\theta$, we compute the Shapley Value $\psi^{k}(f_{\theta}(\mathbb{X}_i))$ of a part $k$ as follows:
\begin{equation}
\label{eq:PSV}
 \psi^{k}(f_\theta(\mathbb{X}_i)) = \sum_{\mathbb{S} \subseteq \mathbb{P} \setminus \{k\}} \frac{|\mathbb{S}|! (K - |\mathbb{S}| - 1)!}{K!} \left[ f_{\theta}(\mathbf{x}_i^{\mathbb{S} \cup \{k\}}) - f_{\theta}(\mathbf{x}_i^{\mathbb{S}}) \right].
\end{equation}
Here, $\mathbb{P}$ is a set of all parts of our interest, \eg in \figref{method} (a), 
$\mathbb{P}=\{$`Hair', `Eye', `Nose', `Mouse', `Ear', `Hand', `Foot'$\}$.
$\mathbb{S}$ is a subset of $\mathbb{P}$, and $\mathbf{x}_i^{\mathbb{S}}$ is a variant of $\mathbf{x}_i$ that only contains parts belonging to $\mathbb{S}$.
Essentially, \eqnref{PSV} is the expectation of the difference between i) a logit vector of a model when the input image contains a certain part $k$ and ii) a logit vector of a model when the input image does not contain the part $k$.
In other words, \eqnref{PSV} indicates the average marginal contribution of a part $k$ across all possible part combinations. 

\begin{figure}[t]
    \centering
    \includegraphics[width=0.9\linewidth]{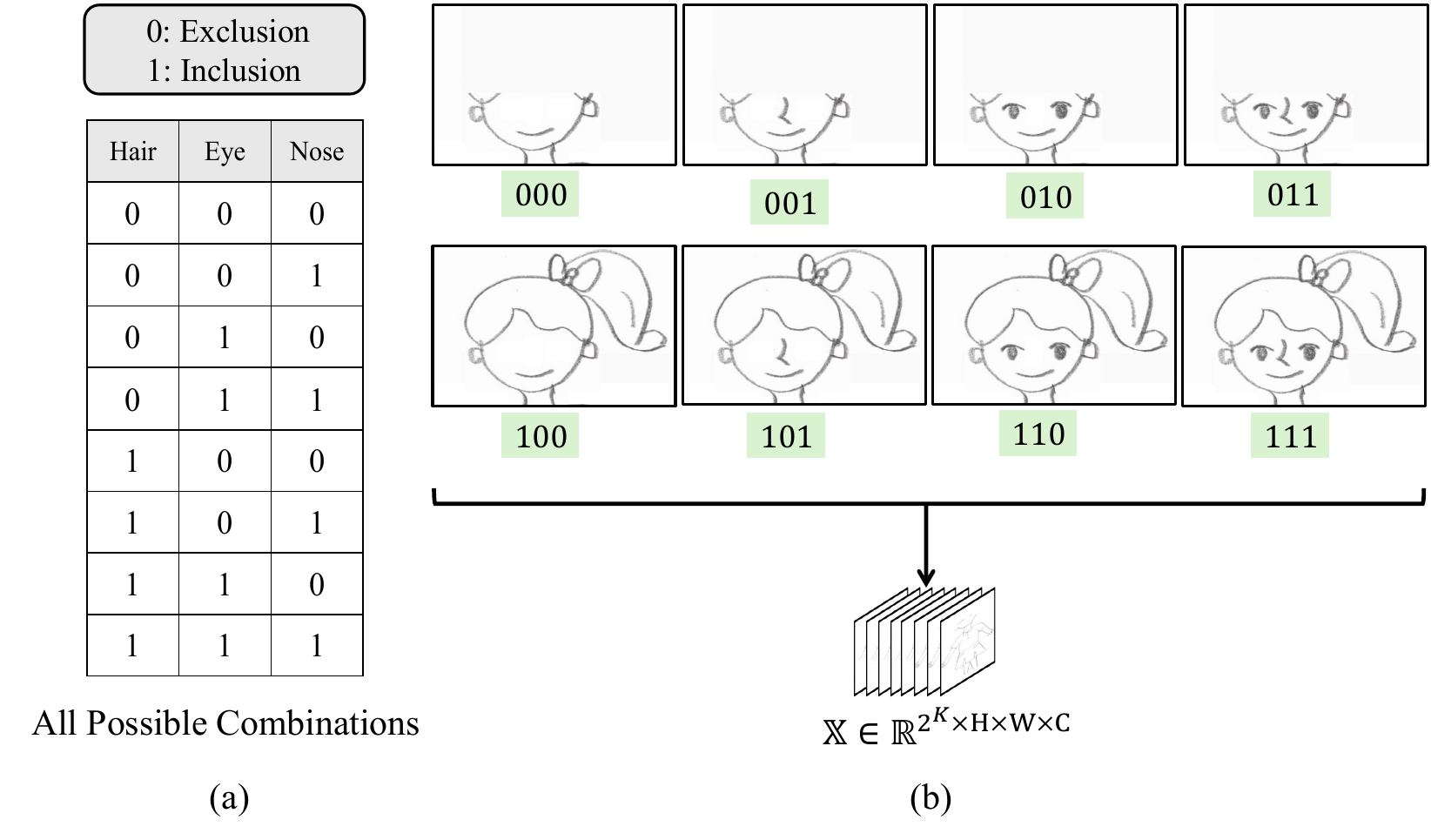}
    \figcaption{Part combination image set generation process}{
    We illustrate the process for generating $\mathbb{X}$, a comprehensive image set consisting of all possible combinations of parts.
    For clarity, we assume only three parts are of interest ($K=3$): `Hair', `Eye', and `Nose'.
    (a) The table shows all eight (\ie $2^K=8$) potential combinations derived from either including (`1') or excluding (`0') each of the three parts. 
    (b) We visualize all part combination images to showcase how each image varies based on the presence or absence of specific parts.
    For instance, an image annotated as `101' contains `Hair' and `Nose' but omits `Eye'. 
    For the omitted part region, we fill in the average pixel value of the input image.
    The collection of the eight generated images forms the set $\mathbb{X}$. 
    }    
    \label{fig:sampling}
\end{figure}

\topic{Part combination image set.}
To compute the Shapley Value $\psi^{k}(f_{\theta}(\mathbb{X}_i))$ using \eqnref{PSV}, we need to prepare the input $\mathbb{X}_i$, \ie all possible combinations of parts given an input image $\mathbf{x}_i$.
In \figref{sampling}, we illustrate how we prepare $\mathbb{X}_i$.
%
Given an input image $\mathbf{x}_i$, we run an off-the-shelf part detector to obtain bounding box coordinates of every part in $\mathbb{P}$.
For the datasets that provide ground-truth part annotations, we use the ground-truth.
Then we mask out each part according to the the combinations.
For example, if we want to generate a part combination, $\mathbf{x}_i^{\{\text{`Hair'}, \text{`Nose'}\}}\in \mathbb{X}_i$, we mask out `Eye' as shown in \figref{sampling}, example `101'.
For masking, we simply inpaint the masked-out region with the average pixel value of the original sample.
Finally, we obtain a set of all possible part combinations $\mathbb{X}_i \in \mathbb{R}^{2^{K} \times H \times W \times C}$.

We can collect the Shapely Values of all $K$ parts to obtain a sample-level part contribution histogram: $\mathbf{c}_{S}=[\psi^1,\psi^2,\ldots,\psi^{K}]^\top$.
Optionally, we can obtain the most contributing part of $\mathbb{X}_i$ by
\begin{equation}
\label{eq:mostpart}
k_i^* = \argmax_k (\mathbf{c}_{S}).
\end{equation}

\subsection{Class-Level and Task-Level \method{}}
\label{sec:group}
We expand our framework from individual sample-level \method{} framework to a broader scope: class-level and task-level \method{} frameworks.
We aggregate all the sample-level part contribution histograms 
$\mathbf{c}_S$
to obtain class-level and task-level statistics.

\topic{C-\method{}.}
The C-\method{} quantifies the contribution of each part within a specific class.
Let us consider a set $\mathbb{D}_c\in\mathbb{D}$ containing $N_c$ samples of the class $c$.
The C-\method{} calculates the most contributing part of every sample in $\mathbb{D}_c$ by \eqnref{mostpart}. 
Then, the C-\method{} constructs a class-level part contribution histogram by counting the most contributing parts as follows:
\begin{align}
    I_k &= \frac{1}{N_c}\sum_{i=1}^{N_c} \mathds{1}(k_i^* = k), \\
    \mathbf{I}_c &= [I_1, I_2, \ldots, I_{K}]^{\top}.
    \label{eq:cpceve}
\end{align}
Here, $I_k$ denotes the frequency at which the $k$-th part is identified as the most significant contributor.
The indicator function, denoted by $\mathds{1}(\cdot)$ outputs 1 when the condition is satisfied and outputs 0 otherwise. 
With \eqnref{cpceve}, the C-\method{} can evaluate the contribution of each part at the class-level.
For instance, in \figref{method} (b), `Hair' is crucial to a model in classifying images as `Female' within the dataset.
In other words, we can explain that the model focuses on `Hair' the most among the seven parts to predict an input image as `Female' on average. 
C-\method{} enhances the interpretability of a model and aids in identifying the importance of each part in decisions related to a specific class.

\topic{T-\method{}.}
The T-\method{} framework extends the C-\method{} to the entire dataset $\mathbb{D}$, combining class-level part contribution histograms, $\mathbf{I}_c$, across all $C$ classes to provide a task-level statistics as follows:
\begin{equation}
    \mathbf{I}_T = \sum_{c=1}^{C} \mathbf{I}_c.
    \label{eq:tpceve}
\end{equation}
The T-\method{} framework evaluates the relative importance of each part in a classification task with a dataset $\mathbb{D}$, offering insights into overall model behavior.
For example, in \figref{method} (c), `Hair' is the most contributing part, and `Foot' is the second most contributing part among all seven parts in a model in distinguishing `Male' and `Female' classes.

In summary, the C-\method{} and T-\method{} frameworks enrich the interpretability of classification models by offering insights into how models prioritize different components in their decision-making processes.
To the best of our knowledge, the \method{} is the \emph{first} approach providing such group-level explanation of classification models.

\section{Results}
\label{sec:results}
In this section, we conduct extensive experiments across various HFD assessment datasets with diverse models to validate the effectiveness of our model explanation framework, \method{}.
Through the experiments, we answer the following research questions: (1) Does the \method{} give reasonable part-based explanations in various HFD assessment tasks? (\secref{sample-level}) (2) Is the \method{} able to provide more abstract level explanations, \ie class-level and task-level? (\secref{group-level}) (3) Is the \method{} able to provide explanations of multiple models? (\secref{other-models}) (4) Can we apply the \method{} to photo-realistic fine-grained classification tasks? (\secref{fine-grained}) To this end, we first provide the details on the dataset used, and our implementation in \secref{datasets} and \secref{impl-details}, respectively.

\subsection{Datasets}
\label{sec:datasets}
%
We evaluate the \method{} using two human figure drawing assessment datasets: the Autism Spectrum Disorder (ASD) screening and the Sketch for Child Art Therapy (SCAT). 
For the extension to the photo-realistic fine-grained visual classification task, we evaluate the \method{} on the Stanford Cars\cite{krause2013car} dataset.

\topic{ASD Screening.}
The ASD Screening~\cite{jongmin2022autism} dataset comprises 100 sketches of human figures, created by subjects diagnosed with autism spectrum disorder (ASD) as well as typically developing (TD) children. 
These subjects range in age from 5 to 12 years. 
In each drawing session, the subject draws a human figure using a pencil and paper.
Each subject draws at least one sketch of the same gender and one of the opposite gender. 
Following the drawing session, the sketches are scanned to obtain digital images.
The task is a binary classification task to distinguish ASD and TD subjects by looking into their drawings.
A model needs to understand subtle differences between drawings drawn by ASD and TD subjects, \eg ASD subjects tend to overemphasize fingers while TD subjects do not.
Due to the data scarcity, we use a 5-fold cross-validation protocol to rigorously evaluate our approach on this dataset.
We ensure all sketches from the same subject belong to the same fold for a fair evaluation. 
Due to confidentiality and the sensitive nature of the data, the dataset is not available for public access~\footnote{All the ASD/TD images visualized in this paper are fake due to the confidentiality issue.}.

\topic{SCAT.}
The SCAT is a public\footnote{Publicly available in the Republic of Korea only.} dataset consisting of drawings for the Human-Tree-Person (HTP) test contributed by child participants.
We use 28,000 human figure drawings in our experiments.
Each participant, from 7 to 13 years old, draws a human figure with a pencil on a piece of paper.
Notably, each drawing includes the gender annotations of the depicted object (drawing) as well as the subject him/herself (drawer).
There are two tasks in this dataset: i) the \emph{drawing} gender classification task, denoted as SCAT-Drawing, and ii) the \emph{drawer} gender classification task, denoted as SCAT-Drawer.
We split the dataset into train and test sets with a ratio of 9:1 for both tasks.

\topic{Stanford Cars.}
Stanford Cars~\cite{krause2013car} dataset comprises 8,144 training images and 8,041 test images covering 196 real-world car models. 
Following prior works on fine-grained visual categorization~\cite{krause2013car, stark2011finecar}
, we focus on classifying 9 car types: cab, convertible, coupe, hatchback, minivan, sedan, SUV, van, and wagon.
We utilize the bounding-box 
annotation provided from a prior work~\cite{chang2021flamingo}.

\subsection{Implementation Details}
\label{sec:impl-details}

\subsubsection{Training}
To validate the \method{}, we train diverse model, including ResNet~\cite{he2016resnet}, DenseNet~\cite{huang2017densenet}, EfficientNet~\cite{tan2019efficientnet} and ViT~\cite{dosovitskiy2021vit} and choose the best performing model for the explanation.
For the evaluation, we fine-tune ImageNet-1K~\cite{imagenet} pre-trained models on the HFD datasets.

\begin{figure*}[t]
\centering
\small
\mpage{0.09}{\includegraphics[width=\linewidth]{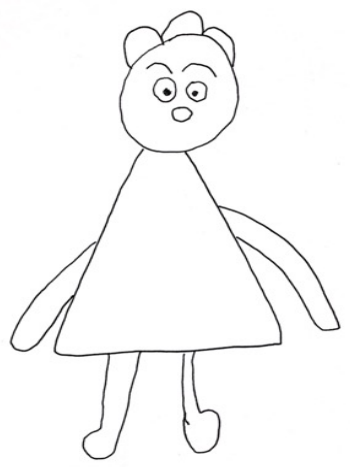}}
\mpage{0.09}{\includegraphics[width=\linewidth]{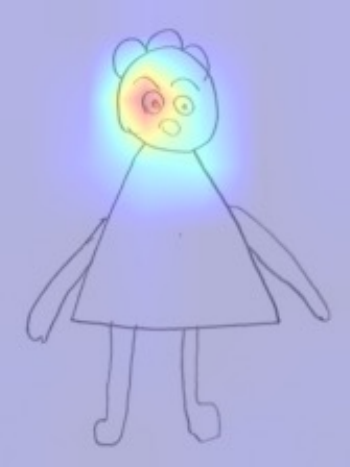}}
\mpage{0.28}{\includegraphics[width=\linewidth]{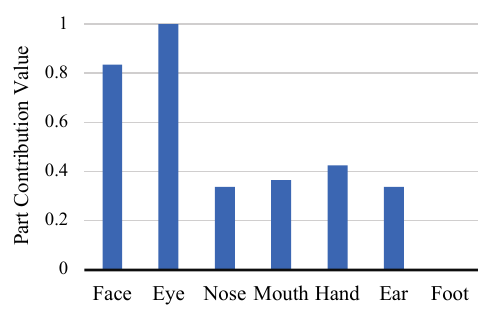}}
\mpage{0.09}{\includegraphics[width=\linewidth]{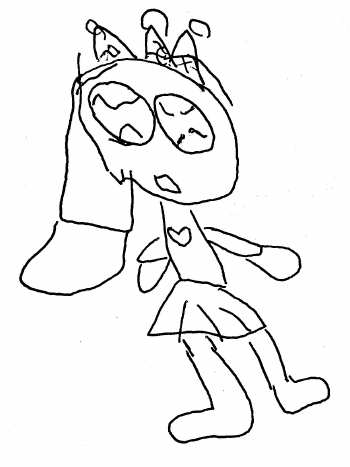}}
\mpage{0.09}{\includegraphics[width=\linewidth]{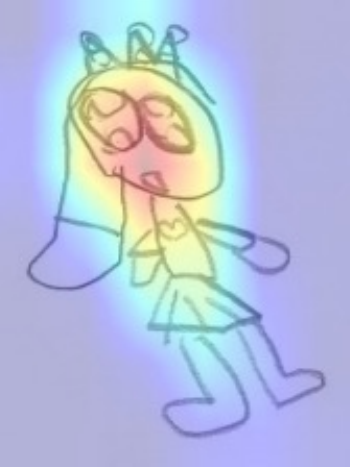}}
\mpage{0.28}{\includegraphics[width=\linewidth]{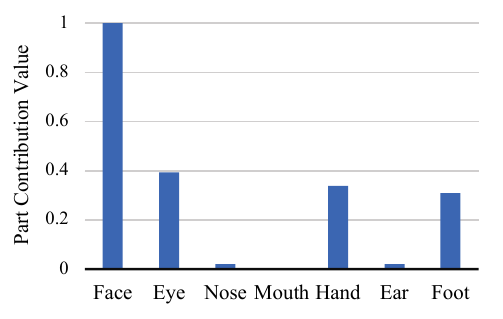}}
\\
\mpage{0.09}{\scriptsize Input}
\mpage{0.09}{\scriptsize GradCAM}
\mpage{0.28}{\scriptsize PCEvE (Ours)}
\mpage{0.09}{\scriptsize Input}
\mpage{0.09}{\scriptsize GradCAM}
\mpage{0.28}{\scriptsize PCEvE (Ours)}
\\
\mpage{0.47}{\scriptsize (a) `ASD' sample of ASD Screening dataset}
\hfill
\mpage{0.47}{\scriptsize (b) `TD' sample of ASD Screening dataset}
\\
\vspace{1em}
\centering
\mpage{0.09}{\includegraphics[width=\linewidth]{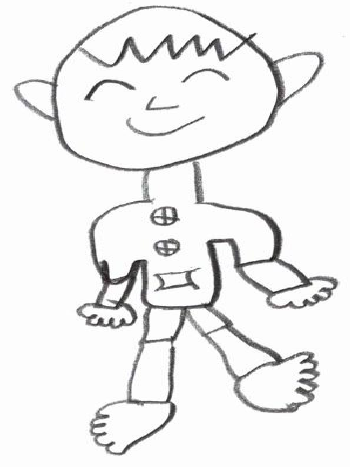}}
\mpage{0.09}{\includegraphics[width=\linewidth]{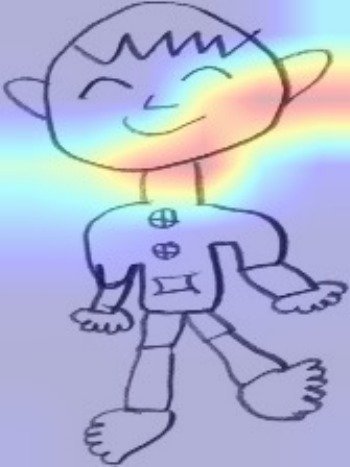}}
\mpage{0.28}{\includegraphics[width=\linewidth]{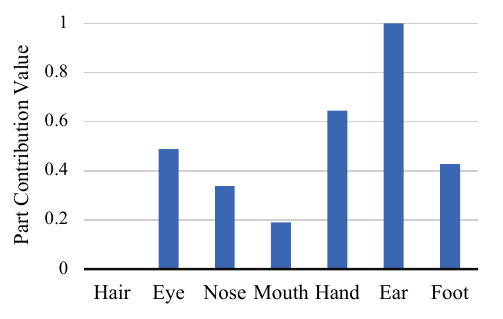}}
\mpage{0.09}{\includegraphics[width=\linewidth]{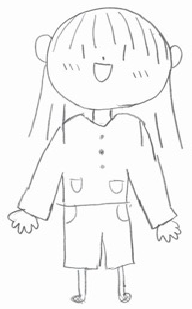}}
\mpage{0.09}{\includegraphics[width=\linewidth]{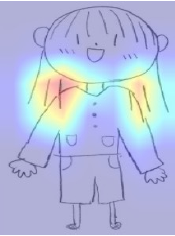}}
\mpage{0.28}{\includegraphics[width=\linewidth]{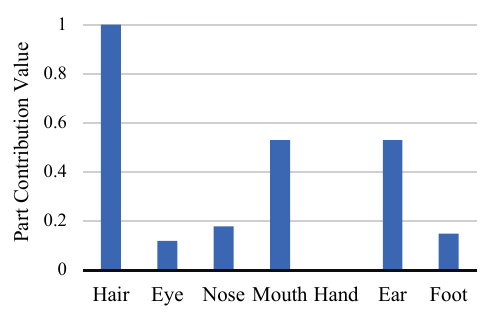}}
\\
\mpage{0.09}{\scriptsize Input}
\mpage{0.09}{\scriptsize GradCAM}
\mpage{0.28}{\scriptsize PCEvE (Ours)}
\mpage{0.09}{\scriptsize Input}
\mpage{0.09}{\scriptsize GradCAM}
\mpage{0.28}{\scriptsize PCEvE (Ours)}
\\
\mpage{0.47}{\scriptsize (c) `Male' sample of SCAT-Drawing dataset}
\hfill
\mpage{0.47}{\scriptsize (d) `Female' sample of SCAT-Drawing dataset}

\caption{\textbf{Sample-level part-based model explanation on the ASD Screening and SCAT datasets.} 
We show model explanations using GradCAM~\cite{selvaraju2017gradcam} and the \method{} on (a) an `ASD' sample and (b) a `TD' sample of the ASD Screening~\cite{jongmin2022autism} dataset, (c) a `Male' sample, and (d) a `Female' sample of the SCAT-Drawing dataset, respectively. 
We normalize each value in the histogram by the maximum value within the sample.
}
\label{fig:sample-level}
\end{figure*}



\begin{table}[t]
\centering
\caption{\tb{Performance comparison of various models on the ASD Screening, SCAT, and Stanford Cars datasets.} We show the top-1 classification accuracy(\%). For the ASD Screening dataset~\cite{jongmin2022autism}, we report the average accuracy and the standard deviation based on results from 5-fold cross-validation. The  \textbf{best} numbers are highlighted.}
\resizebox{\linewidth}{!}{
\begin{tabular}{l cc c ccc c c}
\toprule

\multirow{3}{*}{Model} & \multicolumn{2}{c}{\multirow{2}{*}{ASD Screening~\cite{jongmin2022autism}}}
&& \multicolumn{3}{c}{SCAT} && \multirow{2}{*}{Stanford Cars~\cite{krause2013car}} \\ 
\cline{5-7} 
&&&& Drawing && Drawer \\
\cline{2-3}
\cline{5-5} 
\cline{7-7} 
\cline{9-9}
 & Acc.($\uparrow$) & Std.($\downarrow$) && Acc.($\uparrow$) && Acc.($\uparrow$) && Acc.($\uparrow$) \\ 
\hline
ResNet-18~\cite{he2016resnet} & 89.0 & 6.2 && 96.0 && 73.9 && 89.5 \\
ResNet-50~\cite{he2016resnet} & 89.0 & 5.5 && 95.5 && 73.3 && 90.9\\
DenseNet-121~\cite{huang2017densenet} & 92.0 & 5.7 && \textbf{96.5} && 73.9 && \textbf{94.0}\\
DenseNet-169~\cite{huang2017densenet} & 93.0 & 7.6 && 96.1 && 74.7 && 93.6\\
ViT-T~\cite{dosovitskiy2021vit} & 85.0 & 10.0 && 89.4 && 64.6 && 84.5\\
ViT-S~\cite{dosovitskiy2021vit} & 93.0 & 8.4 && 86.3 && 66.3 && 90.2\\
EfficientNet-B1~\cite{tan2019efficientnet} & \textbf{98.0} & \textbf{2.7} && 95.8 && \textbf{75.5} && 91.8\\ 
\hline

\end{tabular}
}
\label{tab:model_acc}
\end{table}

\topic{Hyperparameters.} For the ASD Screening~\cite{jongmin2022autism} and SCAT datasets, to maintain the aspect ratio of the human object, we resize input images into 224$\times$168 pixels for all models except the ViT. 
For the ViT, we resize the input images into 224$\times$224 pixels to utilize the pre-trained ViT weights.
We normalize input images using the mean and standard deviation calculated from the train set of each HFD dataset. 
The values $\text{(\emph{mean, std.})}$ are (0.975, 0.07) for the ASD Screening~\cite{jongmin2022autism} dataset, and (0.98, 0.065) for the SCAT dataset.
For the Stanford Cars dataset~\cite{krause2013car}, we resize images into 224$\times$224 pixels and use the mean and standard deviation values from the train set of the ImageNet~\cite{imagenet} dataset for normalization.
%
We train models on the ASD Screening~\cite{jongmin2022autism} and SCAT datasets for 30 epochs with a base learning rate of 0.01.
For the Stanford Cars dataset~\cite{krause2013car}, we train for 50 epochs with a base learning rate of 0.001. 
For the evaluation using the ASD Screening dataset~\cite{jongmin2022autism}, we report the average validation accuracy and standard deviation across the 5 folds. 

\topic{Main model selection.} In \tabref{model_acc}, we report the performance of various models on the three datasets. 
Unless we specify the model used, we choose EfficientNet-B1~\cite{tan2019efficientnet} as a main model for evaluating the \method{}, since EfficientNet-B1 shows favorable performance across multiple datasets.

\subsubsection{Part annotations}\label{sec:part-anns}
To apply the \method{}, we need part annotations of each part, \ie bounding boxes. 
The SCAT dataset provides bounding box annotations of 18 parts.
In this work, we use the following seven parts for the evaluation: `eye', `nose', `ear', `mouth', `hand', `foot', and `hair'.

Since the ASD Screening~\cite{jongmin2022autism} dataset does not provide part annotations, we manually annotate the bounding box of each part using the LabelMe toolkit~\cite{russell2008labelme}. 
We annotate nine distinct parts for a sample in the ASD Screening dataset: `head', `eye', `nose', `ear', `mouth', `hand', `foot', `upper body', and `lower body'. 
To provide more clarity, the `head' part includes hair and face, the `upper body' part includes neck and hands, and the `lower body' part includes feet. 
In this work, we use only the following seven parts: `head', `eye', `nose', `ear', `mouth', `hand', `foot', and `face' for the evaluation. 
After carefully observing the samples, we define the `face' part to represent the head area excluding the eyes, nose, mouth, and ears since it also contains discriminative features beneficial in distinguishing ASD and TD samples.
We show some examples in \figref{task-level-ASD-TD} (a) and (b).

Since the Stanford Cars~\cite{krause2013car} dataset also does not provide bounding box annotations, we employ an off-the-shelf part detector, YOLOv3~\cite{redmon2018yolov3} provided by a repository~\footnote{\url{https://github.com/bhadreshpsavani/CarPartsDetectionChallenge}}.
With the part detector, we detect the following five parts in an image: `door', `light', `glass', `sideglass', and `wheel'.

\subsubsection{Inference}
The \method{} can determine which parts the model predominantly focuses on when making predictions. 
To obtain the part statistics, the \method{} needs to infer the single test sample for $2^{K}$ times, where $K$ denotes the number of pre-defined parts.
For the ASD Screening~\cite{jongmin2022autism} dataset inference, we use the model trained on the training set of each fold being validated. 
We use the model checkpoint that achieves the highest validation accuracy for each fold.

\subsection{Sample-level Part-Based Model Explanation}
\label{sec:sample-level}
In \figref{sample-level}, we compare a few explanations of a model \ie EfficientNet-B1~\cite{tan2019efficientnet}, generated by the \method{} and GradCAM~\cite{selvaraju2017gradcam} on the ASD Screening~\cite{jongmin2022autism} dataset and SCAT-Drawing dataset. 
While a pixel-level attribution method, GradCAM, can highlight the region the model of interest focuses on, we still need to interpret the semantic information of the region.
For instance, in \figref{sample-level} (b), GradCAM shows an attention map firing on `face', `eye', `mouth' and `hand'.
To figure out which part contributes the most to the model decision, we, as humans, need to interpret the visualization.
In contrast, the \method{} shows an intuitive part contribution histogram which does not require much interpretation to understand.
Clearly `face' contributes the most and the `eye' contributes the second most to the model decision.
We can observe a similar trend in other examples as well.

\begin{figure}[t]
\centering
\includegraphics[width=.5\linewidth]{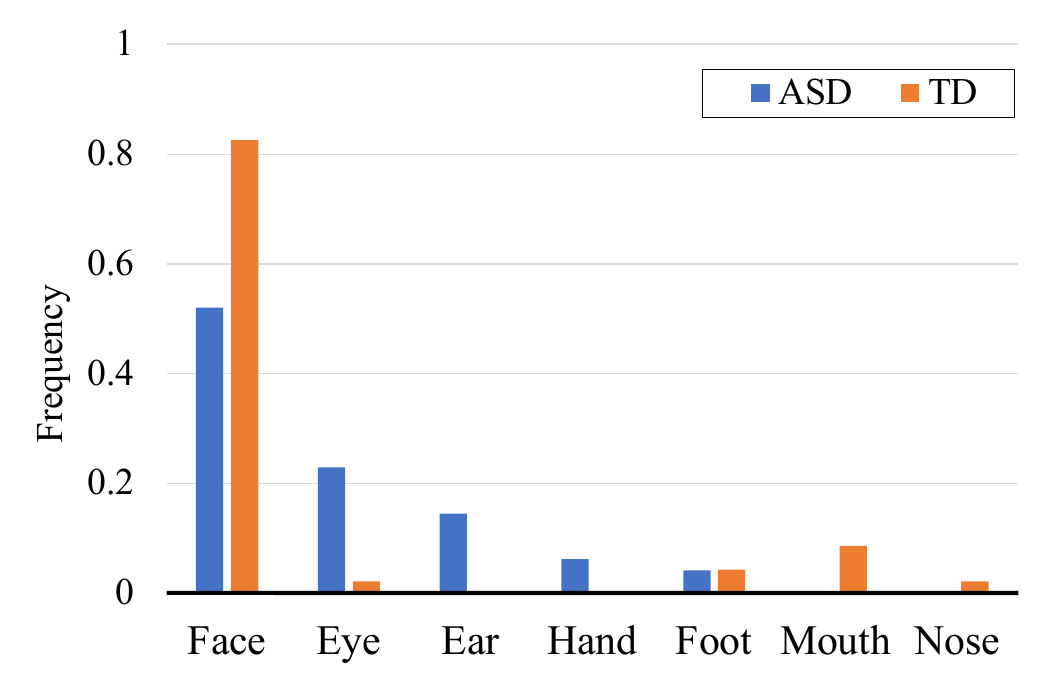}


\caption{\textbf{The class-level PCEvE on the ASD Screening~\cite{jongmin2022autism} dataset.} We show the class-level part contribution histogram generated by the \method{}. 
The histogram shows that `face' contributes the most when the model recognizes ASD and TD samples.
The model tends to focus more on the `face' and less on `ear' and `hand' when recognizing TD samples on average, compared to when recognizing ASD samples.
}
\label{fig:class-level-ASD-TD}
\end{figure}
\subsection{Class-level and Task-level Part-Based Model Explanation}
\label{sec:group-level}
\begin{figure}[t]
\centering
\footnotesize
\mpage{0.55}{
\includegraphics[width=\linewidth]{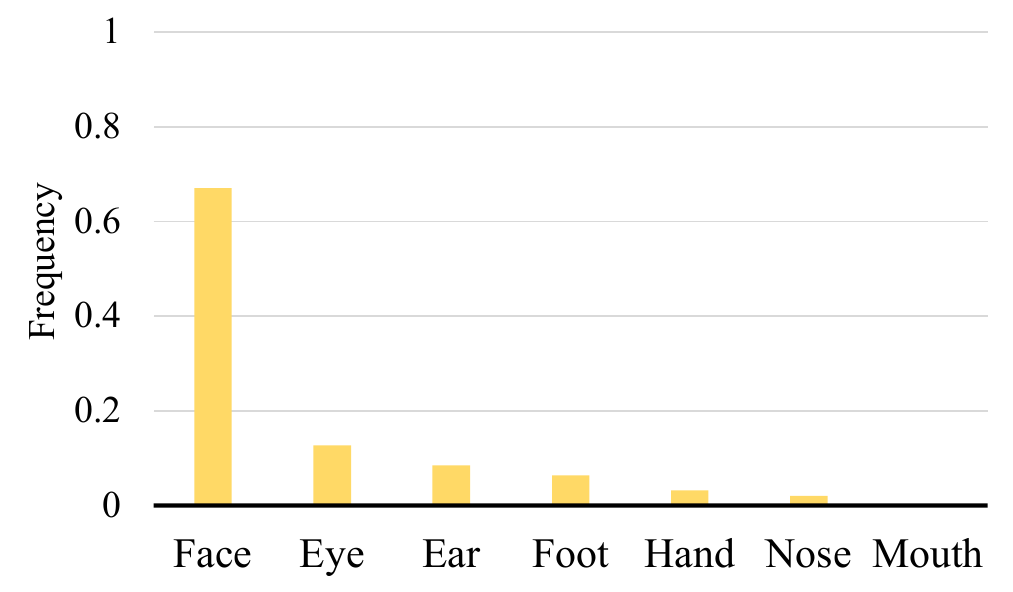}
}
\hfill
\mpage{0.4}{
\includegraphics[width=.9\linewidth]{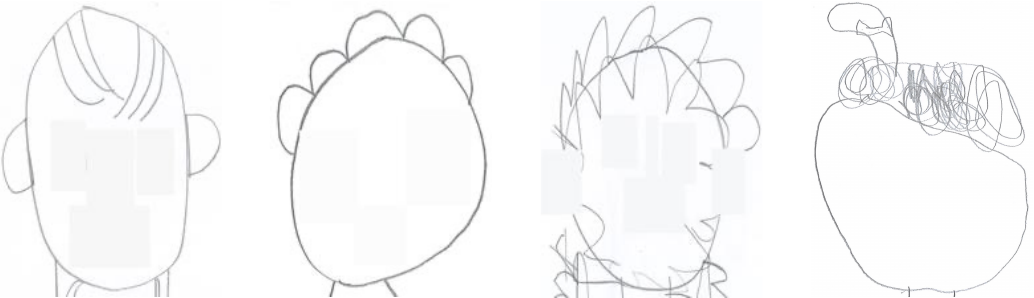}
\\
(a) Face drawn by ASD children
\\
\vspace{2mm}
\includegraphics[width=.9\linewidth]{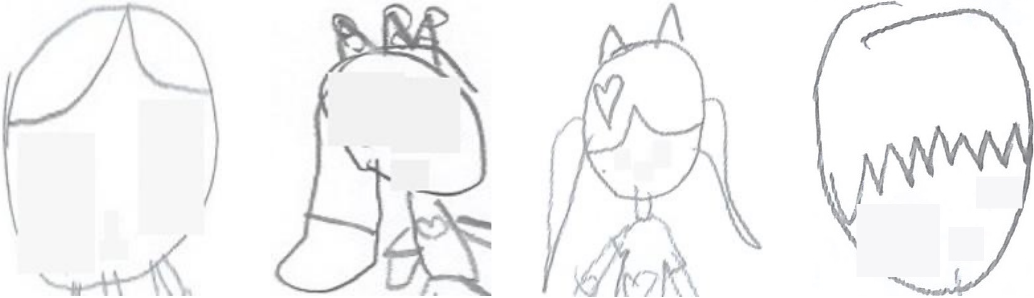}
\\
(b) Face drawn by TD children
}

\caption{\textbf{The task-level PCEvE on the ASD Screening~\cite{jongmin2022autism} dataset.} 
We show the task-level part contribution histogram generated by the \method{}. 
The histogram indicates that the model tends to focus on `face' to distinguish ASD and TD samples.
We also visualize a few examples of `face' drawn by ASD children in (a) and drawn by TD children in (b).
We can see the distinct characteristics of drawings from ASD and TD children.
}
\label{fig:task-level-ASD-TD}
\end{figure}

\topic{ASD Screening.}
We show the class-level and task-level model explanation generated by the \method{} in \figref{class-level-ASD-TD} and \figref{task-level-ASD-TD}, respectively.
When counting the most contributing parts of each sample using \eqnref{cpceve}, we consider the samples correctly predicted by the model only.
In \figref{class-level-ASD-TD}, we observe that the `Face' part is the most contributing part when the model predicts both ASD and TD samples.
The model focuses more on the `Face' and less on the `Ear' and 'Hand' on average when recognizing TD samples, compared to when recognizing ASD samples. 
In \figref{task-level-ASD-TD}, we also observe that the `Face' part is the most contributing part when the model distinguishes ASD and TD samples in the dataset.
As shown in \figref{task-level-ASD-TD} (a) and (b), the `Face' drawn by ASD and TD children show distinct characteristics.

\begin{figure}[t]
\centering
\mpage{0.48}{
\includegraphics[width=\linewidth]{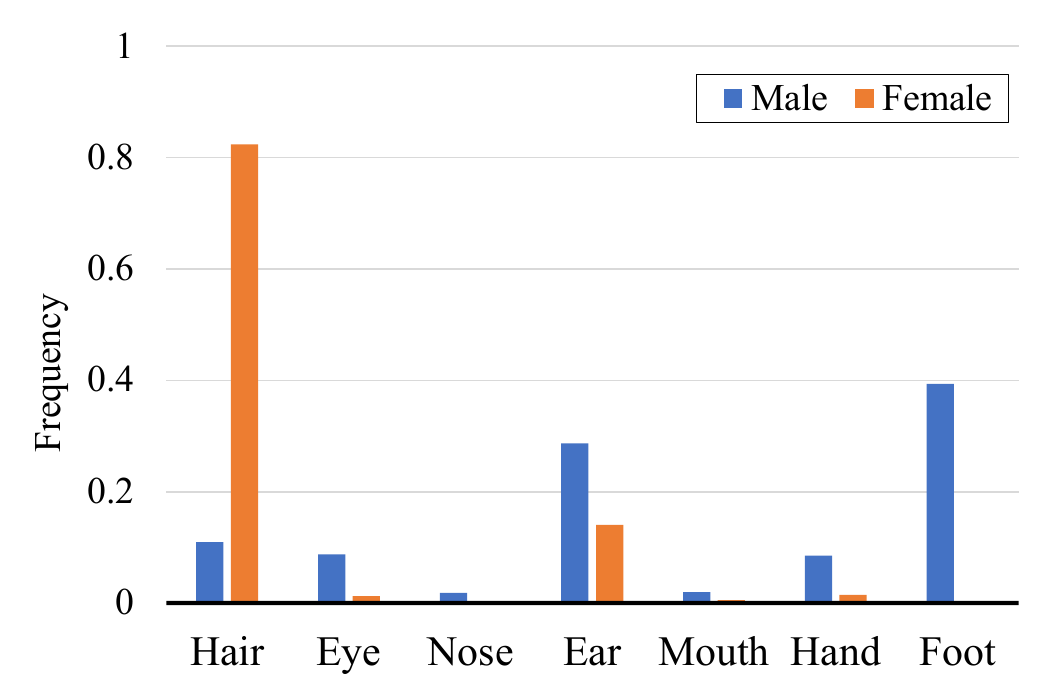}
\\
{\footnotesize (a) Part contribution histogram in the SCAT-Drawing}
\\
}
\mpage{0.48}{
\includegraphics[width=\linewidth]{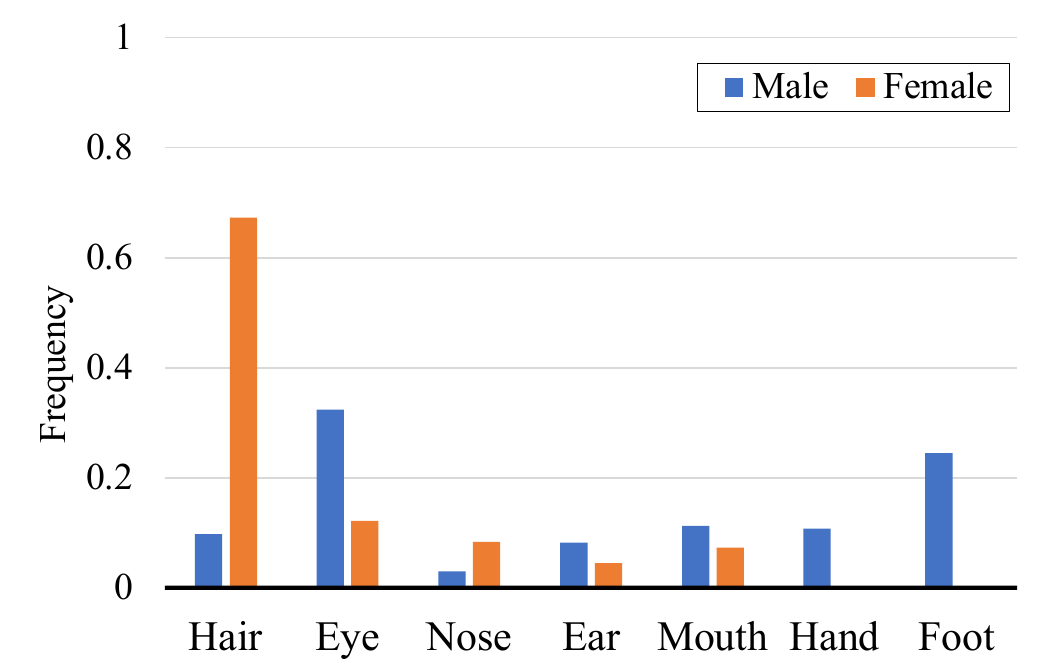}
\\
{\footnotesize (b) Part contribution histogram in the SCAT-Drawer}
}
\caption{\textbf{The class-level \method{} on the SCAT dataset.}
We show the class-level part contribution histogram generated by the class-level \method{}. 
(a) In the SCAT-Drawing, `Foot' turns out to be the most contributing part for the model recognizing the male drawing.
(b) Interestingly, in the SCAT-Drawer, `Eye' turns out to be the most contributing part for the model recognizing the male drawer.
}
\label{fig:class-level-SCAT}
\end{figure}

\begin{figure}[t]
\centering
\footnotesize
\mpage{0.487}{
\includegraphics[width=.9\linewidth]{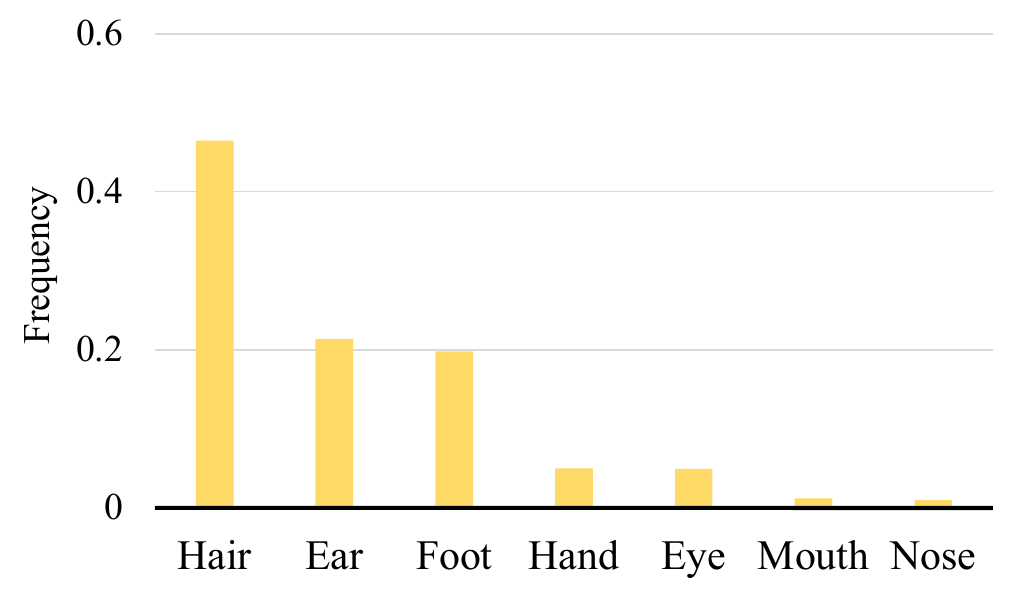}
\\
{\fontsize{7.8pt}{8pt}\selectfont
(a) Part contribution histogram in the SCAT-Drawing
}
\\
\vspace{1em}
\mpage{.45}{
\includegraphics[width=\linewidth]{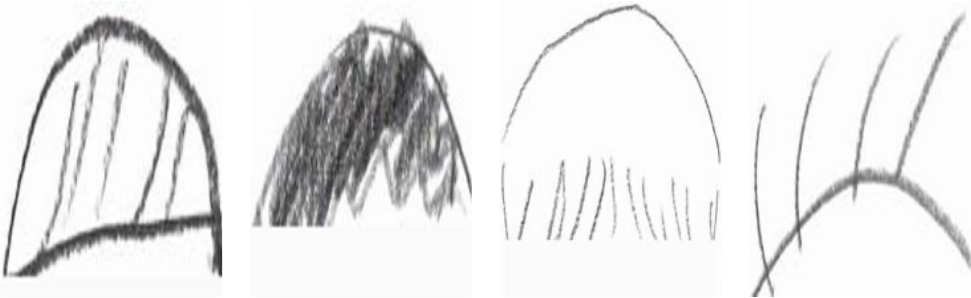}
\\
(c) Hair of male (object) drawing
}
\mpage{.45}{
\includegraphics[width=\linewidth]{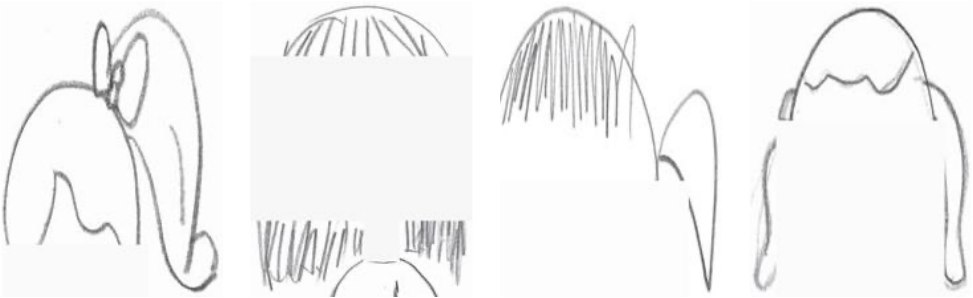}
\\
(e) Hair of female (object) drawing
}
}
\hfill
\mpage{0.487}{
\includegraphics[width=.9\linewidth]{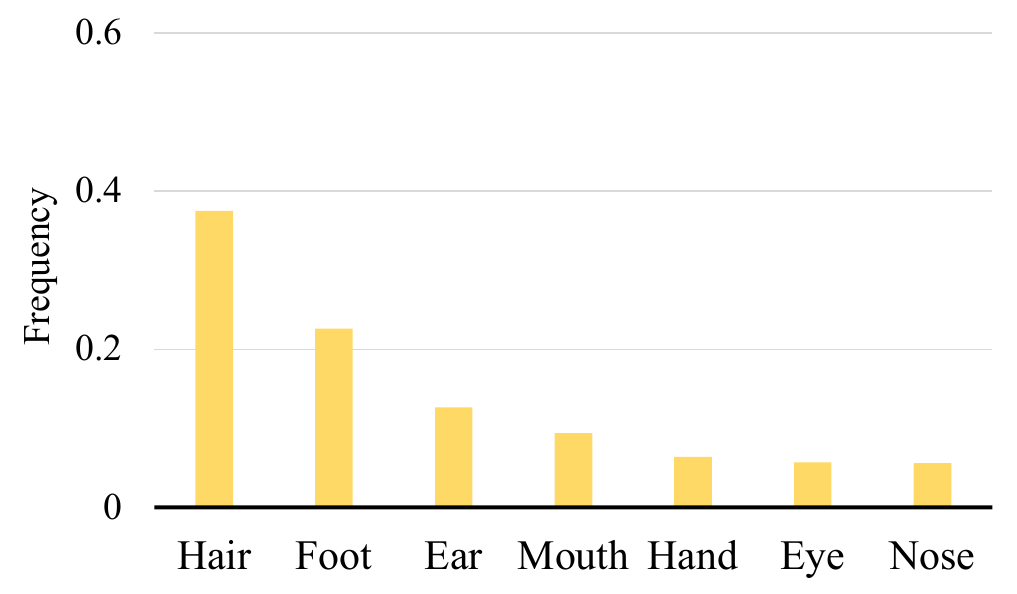}
\\
(b) Part contribution histogram in the SCAT-Drawer
\\
\vspace{1em}
\mpage{.45}{
\includegraphics[width=\linewidth]{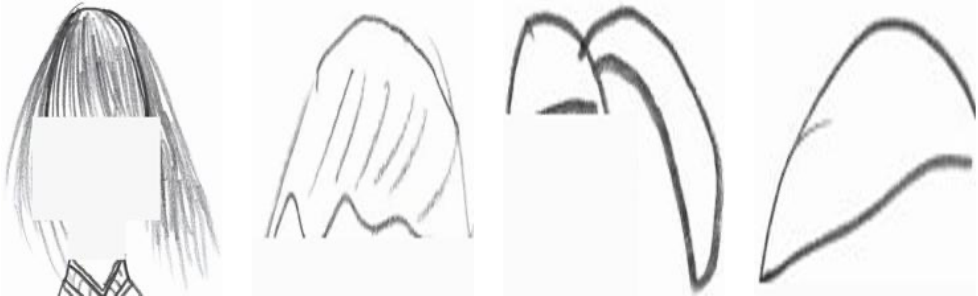}
\\
(d) Hair drawn by male (subject)
}
\mpage{.45}{
\includegraphics[width=\linewidth]{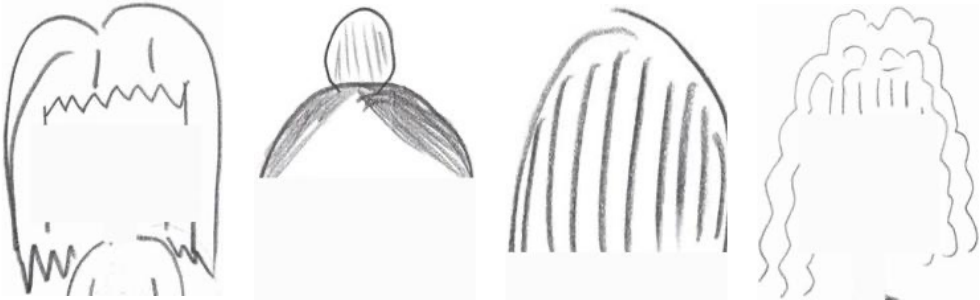}
\\
(f) Hair drawn by female (subject)
}
}

\caption{\textbf{The task-level \method{} on the SCAT dataset.} 
We show the task-level part contribution histogram generated by the \method{}.
Regardless of the task, `Hair' is the most contributing part of the model as shown in (a) and (b).
We also visualize a few examples of `Hair' of male (object) drawing in (c) and `Hair' of female (object) drawing in (e).
We show a few examples of `Hair' drawn by male (subject) in (d) and female (subject) in (f), respectively.
}
\label{fig:task-level-SCAT}
\end{figure}

\topic{SCAT.}
We show the class-level and task-level model explanation generated by the \method{} in \figref{class-level-SCAT} and \figref{task-level-SCAT}, respectively.
When counting the most contributing parts of each sample, we consider the samples correctly predicted by the model only.
In \figref{class-level-SCAT} (a), the class-level part contribution histogram indicates that `Foot' contributes the most for the model recognizing the male drawing while `Hair' contributes the most for the model recognizing the female drawing.
The trend shifts slightly when recognizing the gender of the drawer; here, `Eye' becomes the most contributing part for the model recognizing the male drawer while `Hair' remains same as the most contributing part in recognizing the female drawer as shown in \figref{class-level-SCAT} (b).
We also visualize the task-level part contribution statistics in \figref{task-level-SCAT}. 
We find the `Hair' part is the most discriminative for the model regardless of drawing or drawer gender classification task.
In \figref{task-level-SCAT} (c) and (e), we visualize some sample images of the `Hair' part in each drawing gender class to inspect if there are differences between classes. 
We observe apparent differences in the images of the `Hair' part between drawing object genders.
The `Hair' part can be an important clue even for a human in distinguishing the gender of a human figure drawing, which aligns with the model explanation produced by the \method{}.
In the drawer gender classification task, we observe a similar trend. 
We observe subtle differences between the hair drawn by male and female subjects in \figref{task-level-SCAT} (d) and (f).
%

In summary, the \method{} can provide an abstract part-based explanation of a model.
In the ASD Screening~\cite{jongmin2022autism} and SCAT datasets, the \method{} of the EfficientNet-B1~\cite{tan2019efficientnet} model aligns well with human perception, which might imply the model mimics human perception in the HFD assessment tasks.

\begin{figure}[t]
\centering
\scriptsize

\mpage{0.485}{
\centering
\includegraphics[width=.9\linewidth,height=.115\paperheight]{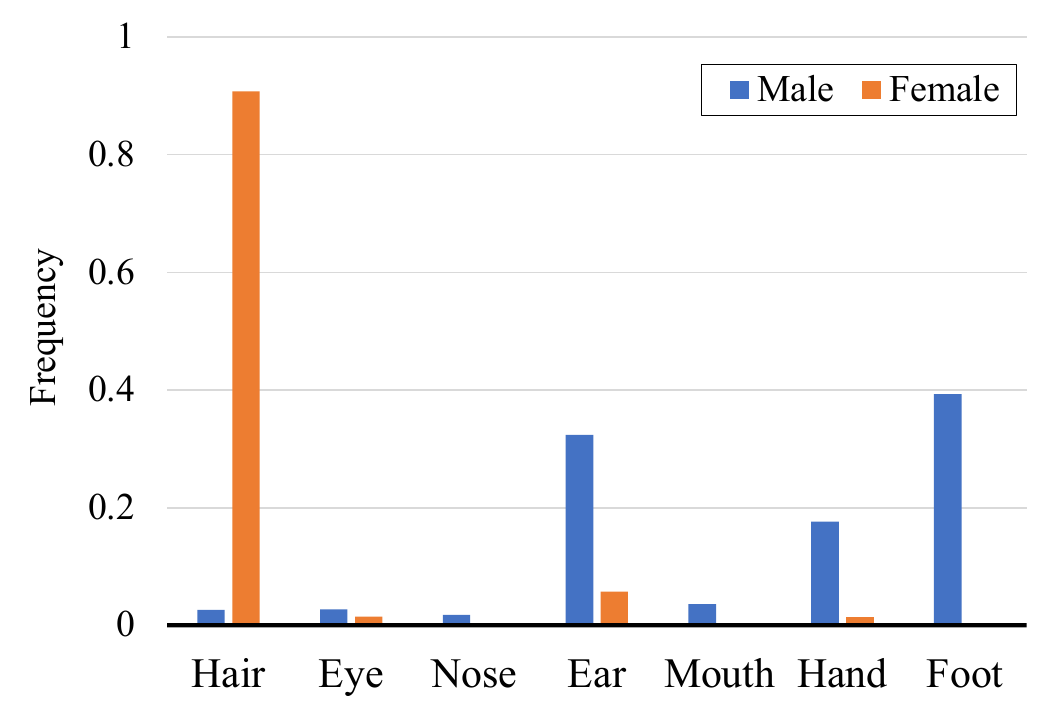}
\\
(a) Drawing gender classification with ViT-Small
}
\hfill
\mpage{0.485}{
\centering
\includegraphics[width=.9\linewidth,height=.115\paperheight]{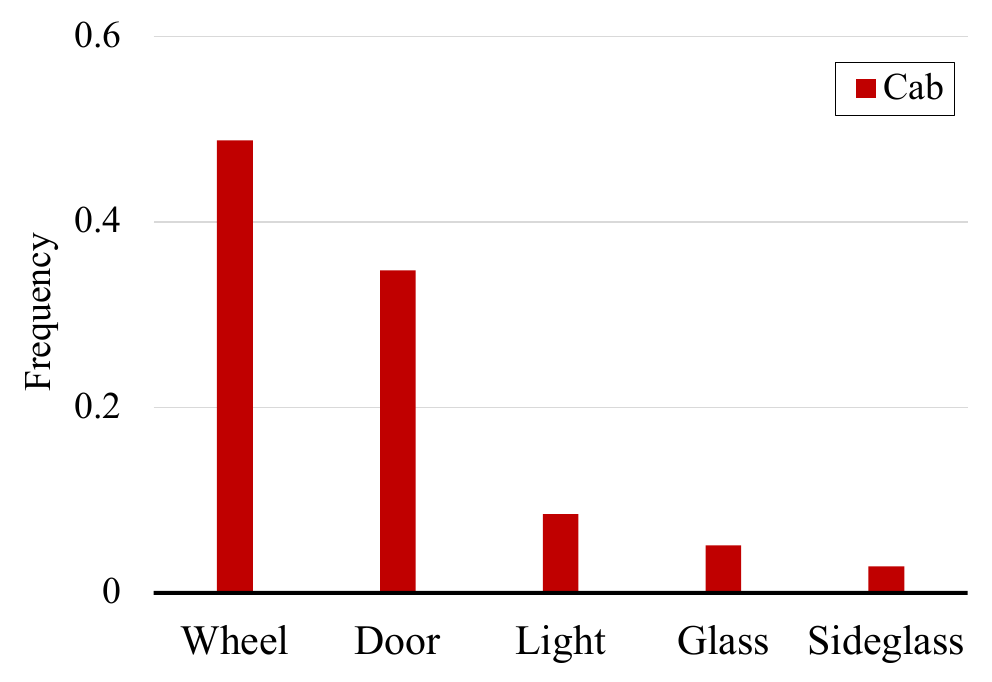}
\\
(b) Car-type classification with ViT-Small
}
\\
\mpage{0.485}{
\centering
\includegraphics[width=.9\linewidth,height=.115\paperheight]{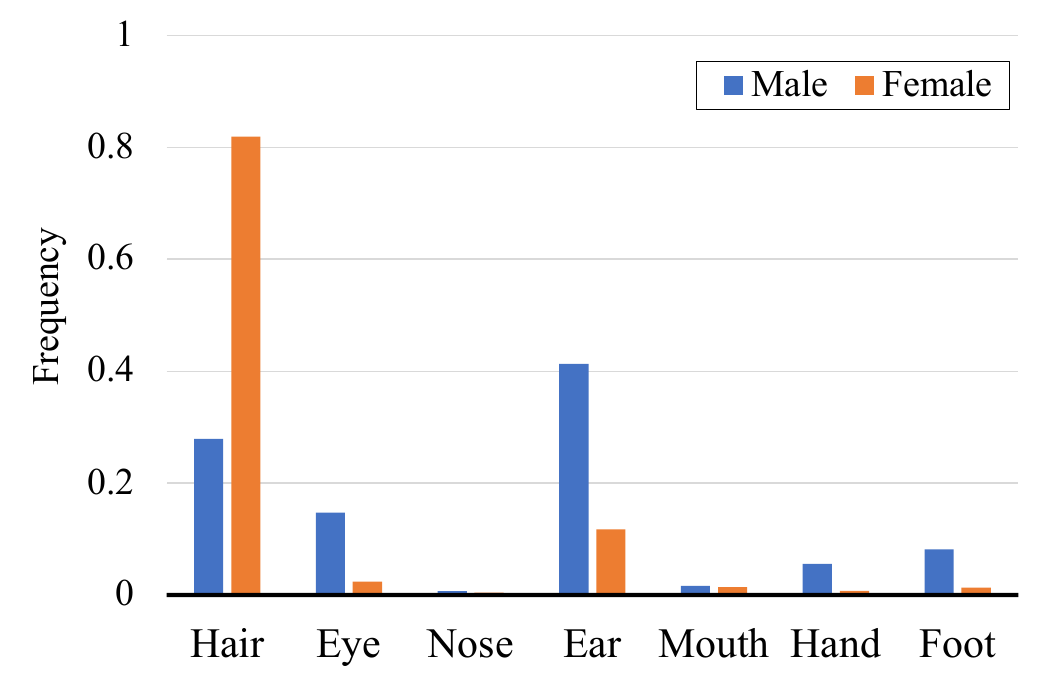}
\\
(c) Drawing gender classification with DenseNet-121
}
\hfill
\mpage{0.485}{
\centering
\includegraphics[width=.9\linewidth,height=.115\paperheight]{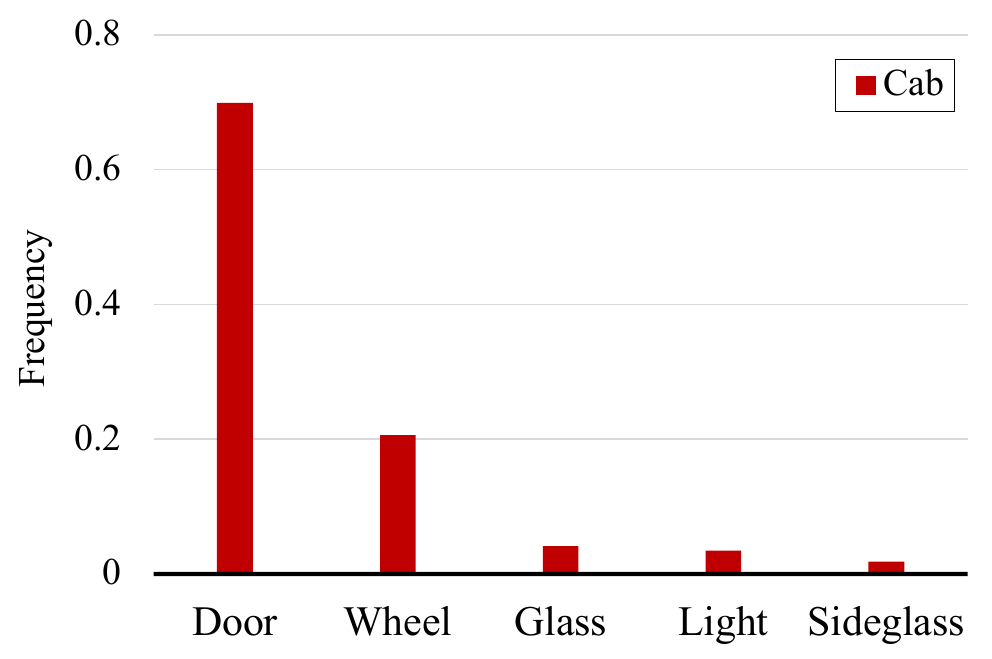}
\\
(d) Car-type classification with DenseNet-121
}


\caption{\textbf{The class-level part contribution evaluation based model explanation of the ViT-Small and the DenseNet-121.}
We show the class-level part contribution histograms generated by the class-level \method{} to explain the ViT-Small~\cite{dosovitskiy2021vit} and the DenseNet-121~\cite{huang2017densenet} on the SCAT-Drawing in (a) and (c) and the Stanford Cars~\cite{krause2013car} in (b) and (d).
For the car-type classification task, we plot the `Cab' class histogram.
%
}

\label{fig:other-models}
\end{figure}

\subsection{Other Model Explanations}
\label{sec:other-models}

In \figref{other-models}, we show the class-level part contribution histograms generated by the \method{} for explaining model behavior across different datasets and models.
Specifically, we examine the ViT-Small~\cite{dosovitskiy2021vit} and DenseNet-121~\cite{huang2017densenet} models on the SCAT-Drawing in \figref{other-models} (a) and (c), and on the Stanford Cars~\cite{krause2013car} dataset in \figref{other-models} (b) and (d). 
For the car-type classification task, we focus on histograms for the 'Cab' class.
In the case of the SCAT-Drawing dataset, both the ViT-Small (\figref{other-models} (a)) and DenseNet-121 (\figref{other-models} (c)) models primarily focus on the `Hair' part for predicting female samples, aligning with observations made for the EfficientNet-B1 model in \figref{class-level-SCAT} (a).
Their focus diverges when classifying male samples: ViT-Small leans towards the `Foot' part, whereas DenseNet-121 prioritizes the `Ear' part for identifying male drawings.

For the car-type classification, the most contributing parts for `Cab' class differ between the two models; ViT-Small focuses on the `Wheel' (in \figref{other-models} (b)), whereas DenseNet-121 focuses on the `Door' (in \figref{other-models} (d)) the most.
In summary, we validate that the \method{} can provide reasonable explanations across multiple models including both CNNs and Transformers.

\subsection{Extension to Photo-realistic Fine-grained Visual Categorization}
\label{sec:fine-grained}
\begin{figure}[t]
\centering
\mpage{0.48}{
\centering
\mpage{0.32}{
\includegraphics[width=\linewidth]{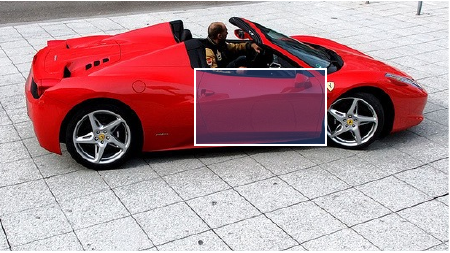}
}
\mpage{0.62}{
\includegraphics[width=\linewidth]{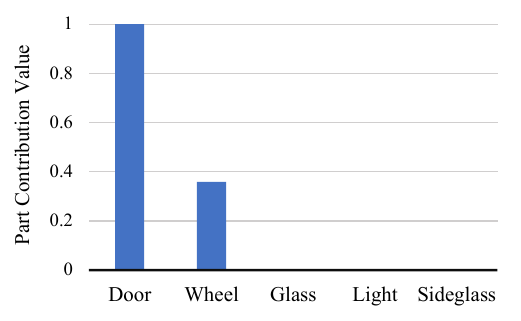}
}
\\
\footnotesize{(a) a `Convertible' sample}
}
\mpage{0.48}{
\centering
\mpage{0.32}{
\includegraphics[width=\linewidth]{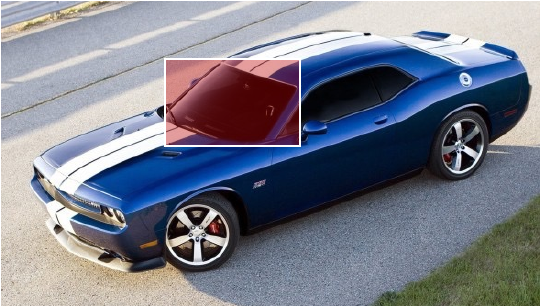}
}
\mpage{0.62}{
\includegraphics[width=\linewidth]{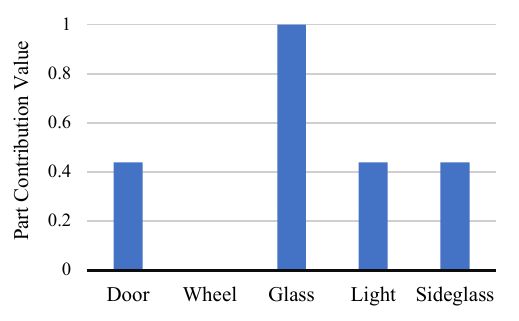}
}
\\
\footnotesize{(b) a `Coupe' sample}
}

\caption{\textbf{Sample-level \method{} on the Stanford Cars~\cite{krause2013car} dataset.} 
(a) The `Door' part contributes the most when the model recognizes the `Convertible' image.
(b) The `Glass' part contributes the most when the model recognizes the `Coupe' image. 
We normalize each value in the histogram by the maximum value within the sample.
}
\label{fig:sample-level-cars}
\end{figure}
\begin{figure}[t]
\centering
\footnotesize 
\mpage{0.31}{
\includegraphics[width=\linewidth]{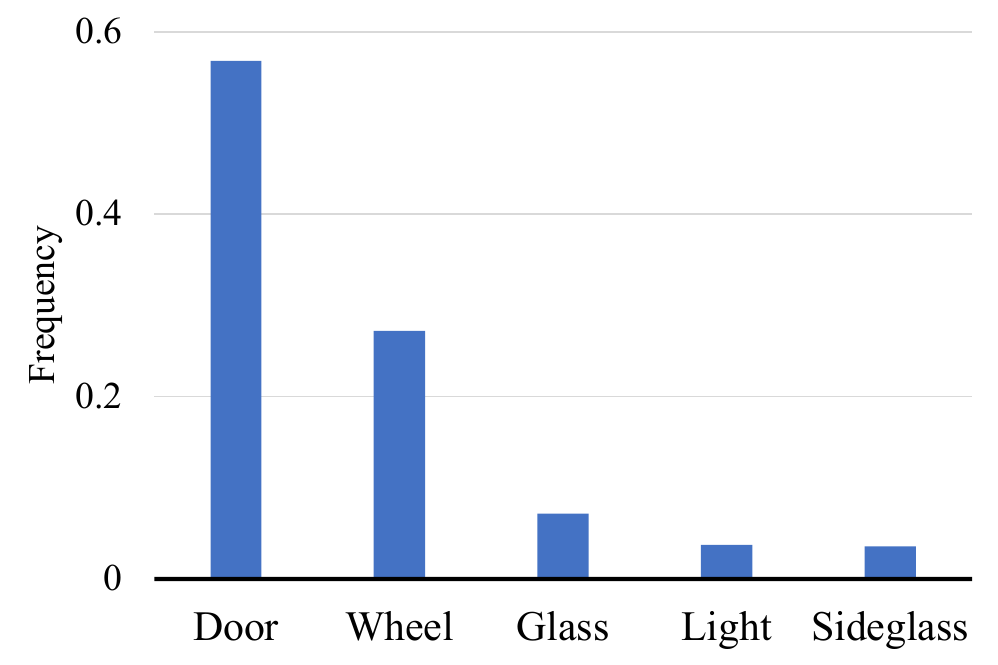}
\\
\centering
(a) The `Cab' class
\\
\vspace{1em}
\includegraphics[width=0.45\linewidth]{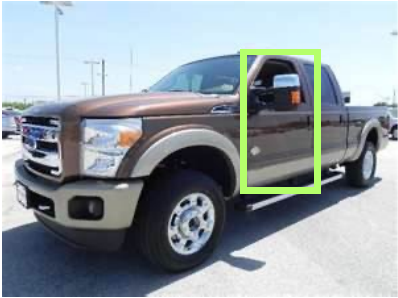}
\includegraphics[width=0.45\linewidth]{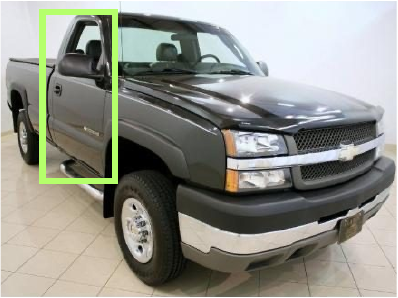}
\\
\mpage{0.45}{\footnotesize Cab}
\mpage{0.45}{\footnotesize Cab}
\\
\vspace{0.5em}
\includegraphics[width=0.45\linewidth]{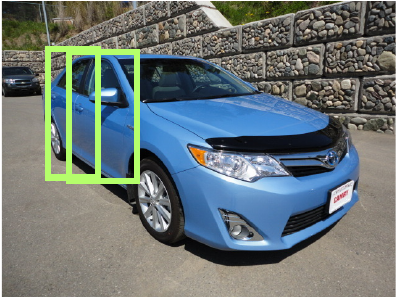}
\includegraphics[width=0.45\linewidth]{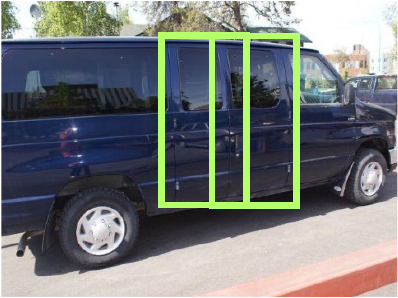}
\\
\mpage{0.45}{\footnotesize Sedan}
\mpage{0.45}{\footnotesize Wagon}
}
\hfill
\mpage{0.31}{
\includegraphics[width=\linewidth]{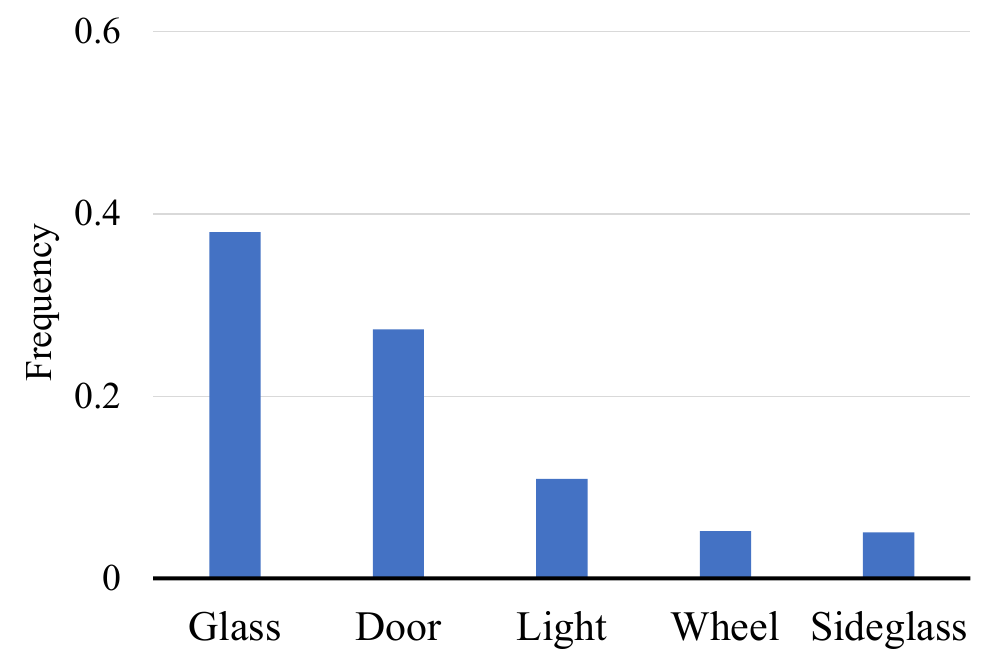}
\\
\centering
(b) The `Coupe' class
\\
\vspace{1em}
\includegraphics[width=0.45\linewidth]{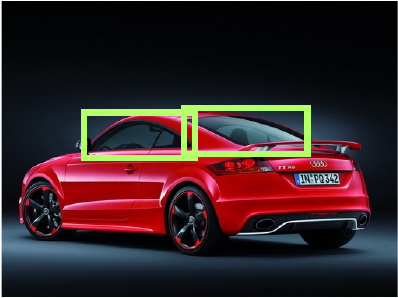}
\includegraphics[width=0.45\linewidth]{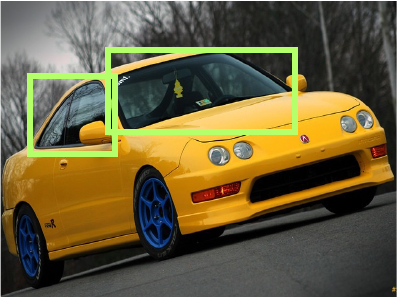}
\\
\mpage{0.45}{\footnotesize Coupe}
\mpage{0.45}{\footnotesize Coupe}
\\
\vspace{0.5em}
\includegraphics[width=0.45\linewidth]{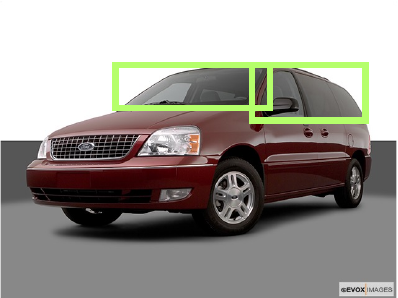}
\includegraphics[width=0.45\linewidth]{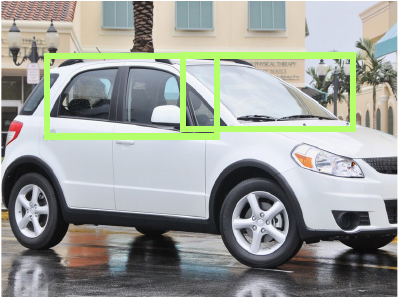}
\\
\mpage{0.45}{\footnotesize Minivan}
\mpage{0.45}{\footnotesize Hatchback}
}
\hfill
\mpage{0.31}{
\includegraphics[width=\linewidth]{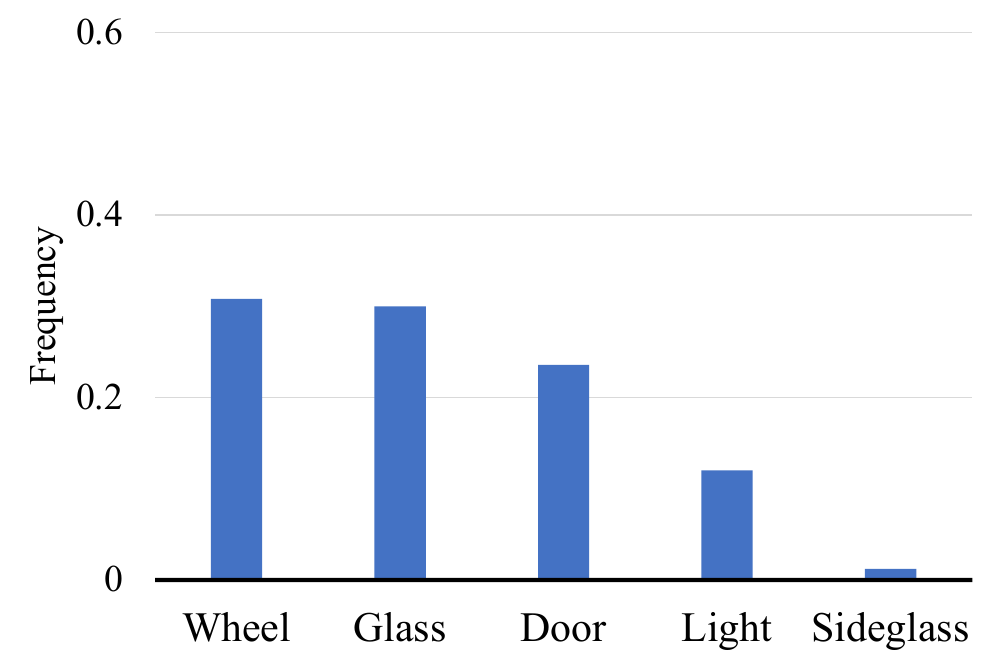}
\\
\centering
(c) The `Van' class
\\
\vspace{1em}
\includegraphics[width=0.45\linewidth]{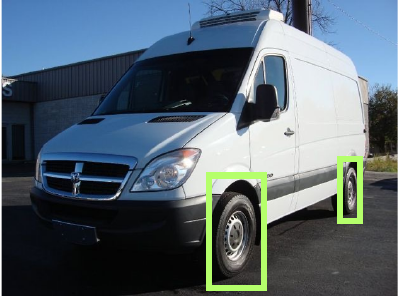}
\includegraphics[width=0.45\linewidth]{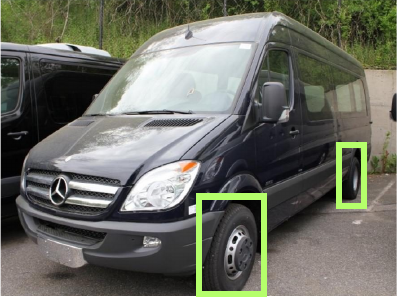}
\\
\mpage{0.45}{\footnotesize Van}
\mpage{0.45}{\footnotesize Van}
\\
\vspace{0.5em}
\includegraphics[width=0.45\linewidth]{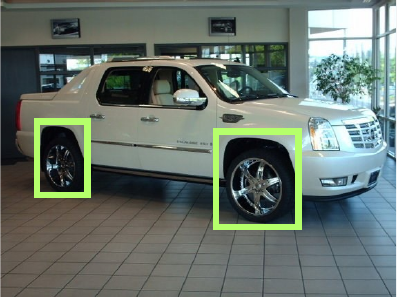}
\includegraphics[width=0.45\linewidth]{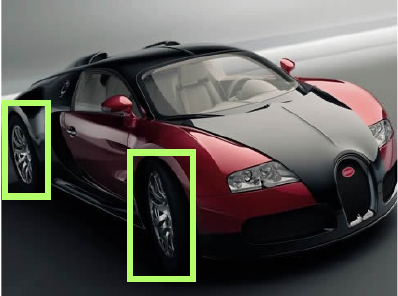}
\\
\mpage{0.45}{\footnotesize Cab}
\mpage{0.45}{\footnotesize Coupe}
}

\caption{\textbf{The class-level \method{} on the Stanford Cars~\cite{krause2013car} dataset.}
We show the class-level part contribution histograms generated by the \method{}: (a) the `Cab' class, (b) the `Coupe' class, and (c) the `Van' class.
We also present samples highlighting the most important part from each class (in the second row) along with samples from other classes (in the third row) for comparison.
}
\label{fig:class-level-cars}
\end{figure}

In this section, we move beyond the human figure drawing assessment tasks.
We showcase that our \method{} framework is also suitable for the photo-realistic Fine-Grained Visual Categorization (FGVC) task. 
The FGVC is a task where the goal is to classify images into find-grained categories, \eg 200 different bird species~\cite{wah2011cub}.
There are several image datasets~\cite{wah2011cub, maji2013fgvcaircraft}
for the FGVC task. 
These datasets share common characteristics: each sample consists of the same composition of parts, \eg every bird has eyes, a beak, and wings, and a model should be able to pick up the subtle visual differences in specific parts that distinguish one class from others.
Therefore, we can validate whether the \method{} shows reasonable part-based explanations of a model or not on an FGVC dataset.
Here, we validate the \method{} framework on the Stanford Cars~\cite{krause2013car} dataset.

\topic{Sample-level \method{}.}
In \figref{sample-level-cars}, the `Door' part contributes the most when the model recognizes the `Convertible' image in (a).
The `Glass' part contributes the most when the model recognizes the `Coupe' image in (b).
We find that the part-based explanations generated by the \method{} align well with human perception.
For instance, the `Glass' part of a `Coupe' shows a sloping rear roofline, which is a common characteristic of coupe cars.
We observe the highest part contribution value of the `Glass' in this example.

\topic{Class-level \method{}.}
We show the class-level model explanation generated by the \method{} and some example car images highlighting the most contributing part in \figref{class-level-cars}.
In \figref{class-level-cars} (a) the `Door' part is the most contributing part in the model recognizing `Cab' images.
With a visual inspection, we find the `Door' part is highly discriminative even for humans in recognizing the `Cab' vehicles from `Sedan' or `Wagon' vehicles.

\subsection{Sanity Check Experiments}
\label{sec:discussion}
In this section, we validate the effectiveness and reliability of the \method{} by addressing several critical research questions.
(1) Does the inclusion or exclusion of the most important part extracted by \emph{class-level} \method{} indeed significantly affect the model predictions? (\secref{partwise-inference})
(2) Is there a substantial difference between the feature spaces of the most contributing part images and the least contributing part images? (\secref{tsne})
(3) Is our method relying on the ground-truth part annotations? (\secref{detector-anns})
We conduct all the experiments using the model trained with original images.
\begin{figure}[!h]
\centering
\scriptsize
\begin{minipage}[t]{0.32\linewidth}
\centering
\includegraphics[width=\linewidth]{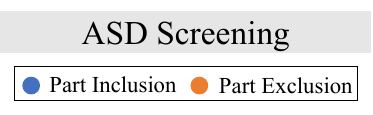}
\\
\includegraphics[width=\linewidth]{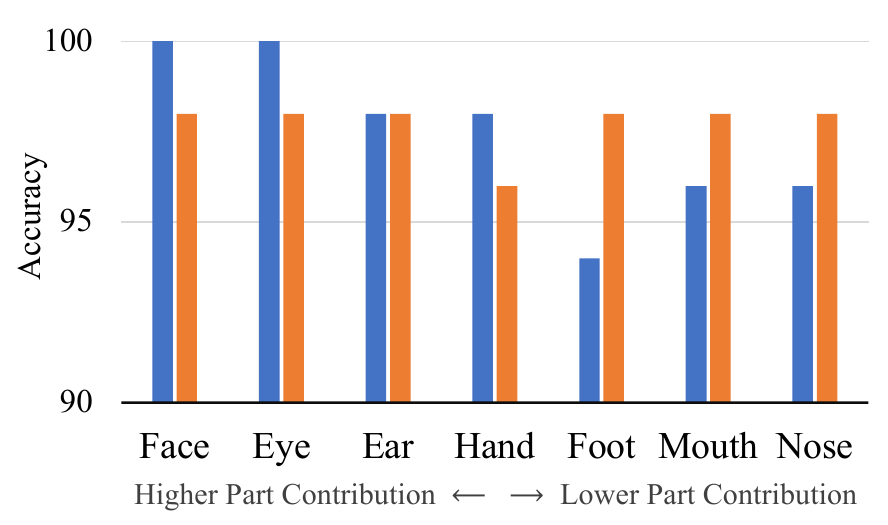}
\\
(a) The `ASD' class
\\
\includegraphics[width=\linewidth]{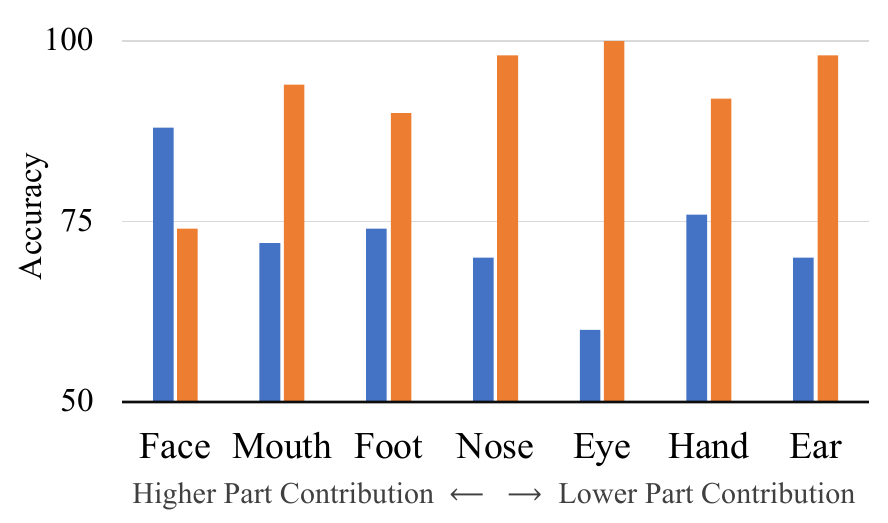}
\\
(b) The `TD' class
\end{minipage}
\begin{minipage}[t]{0.32\linewidth}
\centering
\includegraphics[width=\linewidth]{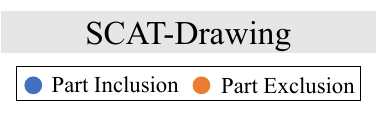}
\\
\includegraphics[width=\linewidth]{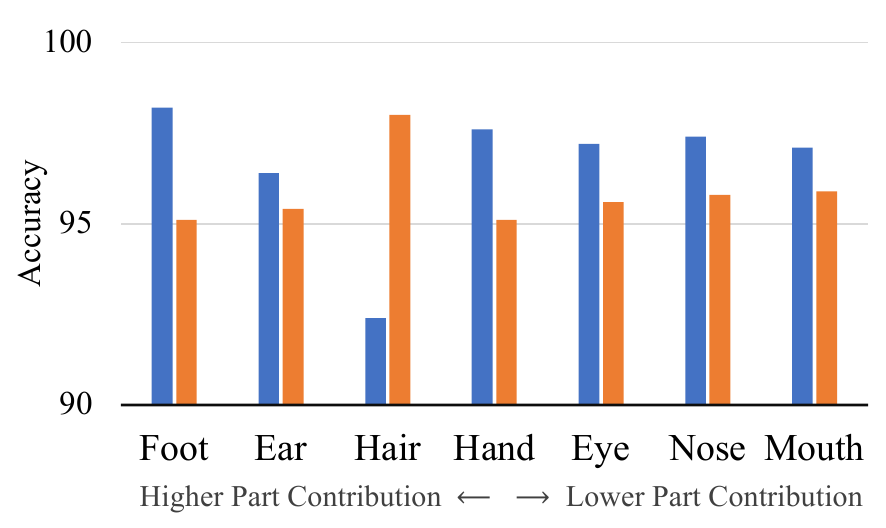}
\\
(c) The `Male' class
\\
\includegraphics[width=\linewidth]{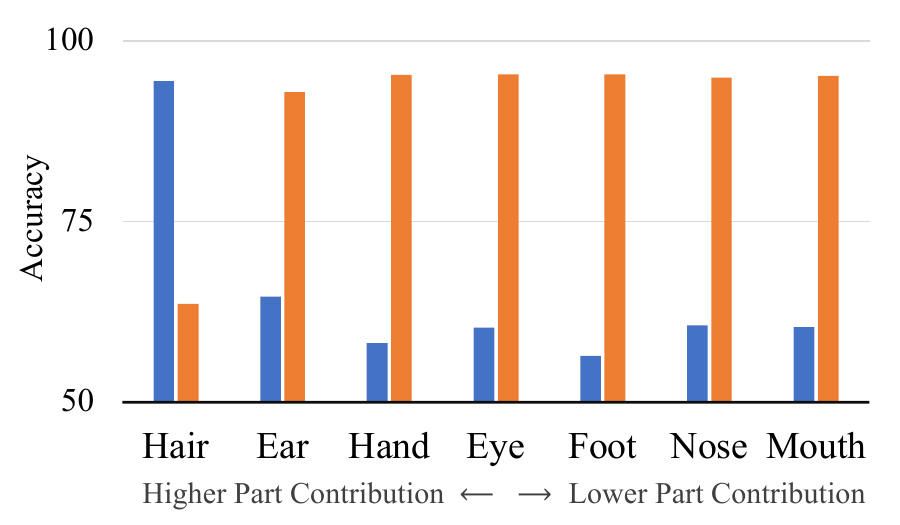}
\\
(d) The `Female' class
\end{minipage}
\begin{minipage}[t]{0.32\linewidth}
\centering
\includegraphics[width=\linewidth]{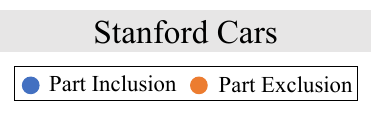}
\\
\includegraphics[width=\linewidth]{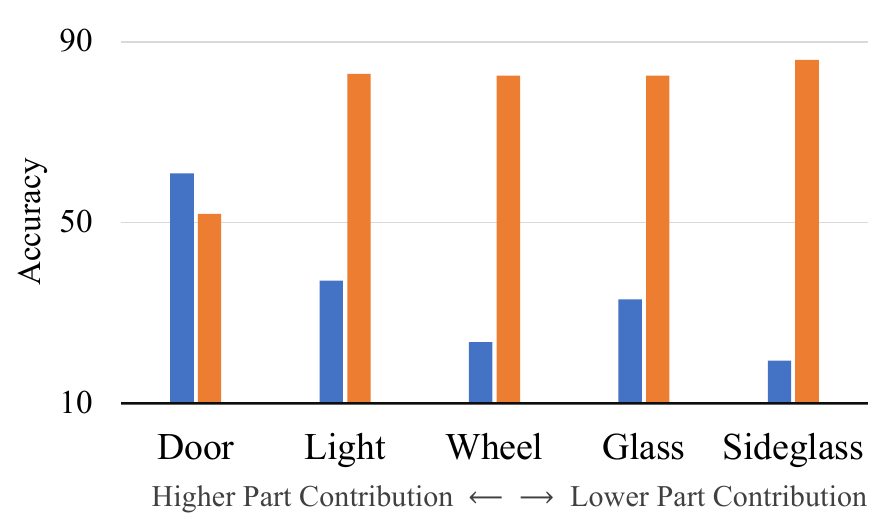}
\\
(e) The `Wagon' class
\\
\includegraphics[width=\linewidth]{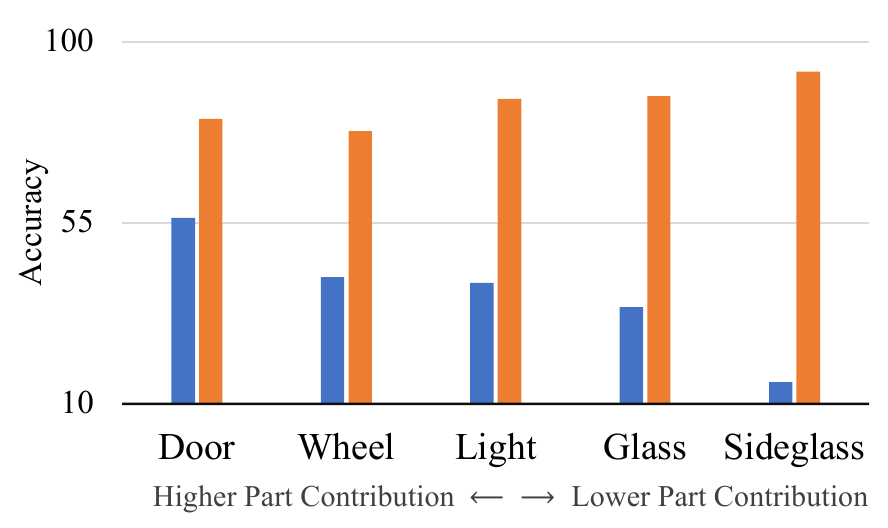}
\\
(f) The `Minivan' class
\end{minipage}


\caption{
\tb{Results of part inclusion and exclusion experiments.} 
We show the results of the part inclusion and exclusion experiments on the ASD Screening~\cite{jongmin2022autism} dataset (first row), the SCAT-Drawing dataset (second row), and the Stanford Cars~\cite{krause2013car} dataset (third row).
In the part inclusion experiments, we assess classification accuracy with inputs that include only each specific part (alongside a torso as the default), represented by \blue{blue} bars.
Conversely, the part exclusion experiments involve inputs that lack each specific part, indicated by \orange{orange} bars. 
Each bar plot is sorted in descending order of part contribution value from left to right.
%
}
\label{fig:partwise-inference}
\end{figure}

\subsubsection{Class-level Validation: Part Inclusion and Exclusion Experiments}
\label{sec:partwise-inference}

We empirically validate the reliability of the class-level \method{} by i) part inclusion and ii) exclusion experiments.
In the part inclusion experiment, we evaluate the classification accuracy using inputs containing each part only (with a torso by default), \eg using \emph{eyes only} or \emph{hair only} with a torso by default for the ASD screening task. 
We expect that using only the most contributing part determined by the \method{} for the prediction shows the highest performance compared to using any other part only for the prediction.
In the part exclusion experiment, we evaluate the classification performance using inputs not containing each part, \eg \emph{not} using \emph{eyes} or \emph{hair} for the ASD screening task. 
The expectation is that not using the most contributing part determined by the \method{} for the prediction shows the lowest performance compared to not using any other part for the prediction.

In \figref{partwise-inference}, we show the result of the part inclusion and exclusion experiments conducted across three datasets: the ASD Screening~\cite{jongmin2022autism} dataset in the first row, the SCAT-Drawing dataset in the second row, and the Stanford Cars~\cite{krause2013car} dataset in the third row.
Each bar plot is sorted in descending order of part contribution value from left to right.
In six out of six cases, we observe the classification with the most contributing part determined by the method{} only shows the highest accuracy: \eg the inclusion of the `Hair' part results in the female drawing recognition accuracy increases more than $50\%p$.
Conversely, the classification without the most contributing part determined by the method{} shows the lowest accuracy in five out of six cases, \eg the exclusion of the `Face' part causes a substantial drop in TD recognition accuracy by $25\%p$.
The results show a clear trend: the inclusion or exclusion of a most/least contributing part determined by the \method{} results in a significant accuracy variation.
The results validate the effectiveness of our method in identifying key parts of input images in a model decision.

\subsubsection{Validation by T-SNE visualization with part features}\label{sec:tsne}
\begin{figure}[t]
\centering
\scriptsize
\begin{minipage}[t]{0.32\linewidth}
\centering
\vspace{0pt}
\includegraphics[width=\linewidth]{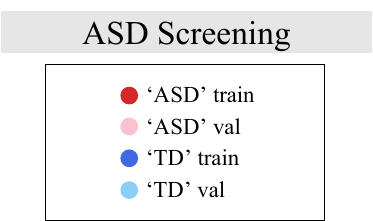}
\\
\vspace{0.05cm}
\includegraphics[width=\linewidth]{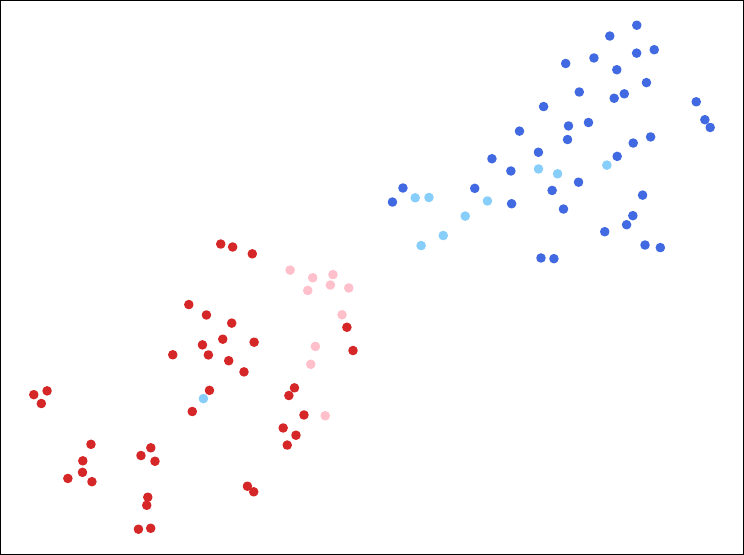}
\\
(a) `Face' part only \\(most contributing)
\\
\includegraphics[width=\linewidth]{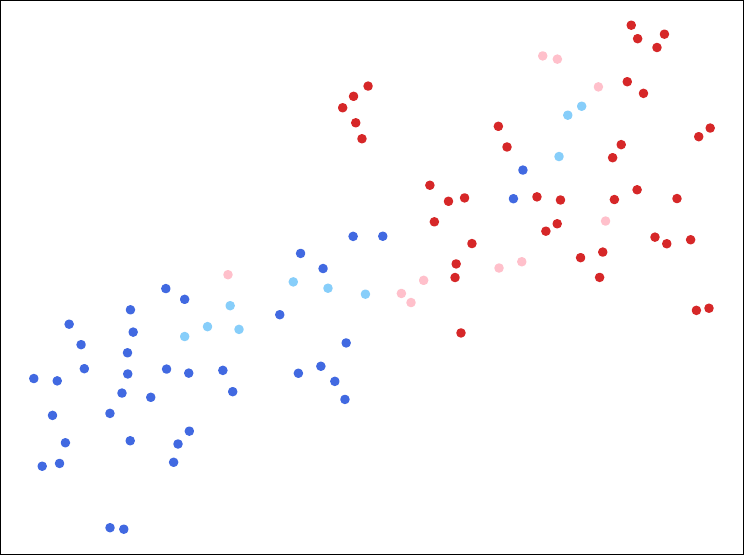}
\\
(b) `Mouth' part only \\(least contributing)
\end{minipage}
\begin{minipage}[t]{0.32\linewidth}
\centering
\vspace{0pt}
\includegraphics[width=\linewidth]{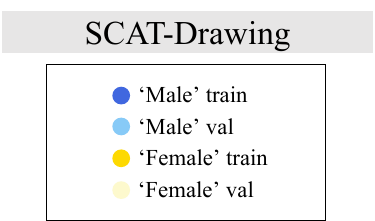}
\\
\vspace{0.05cm}
\includegraphics[width=\linewidth]{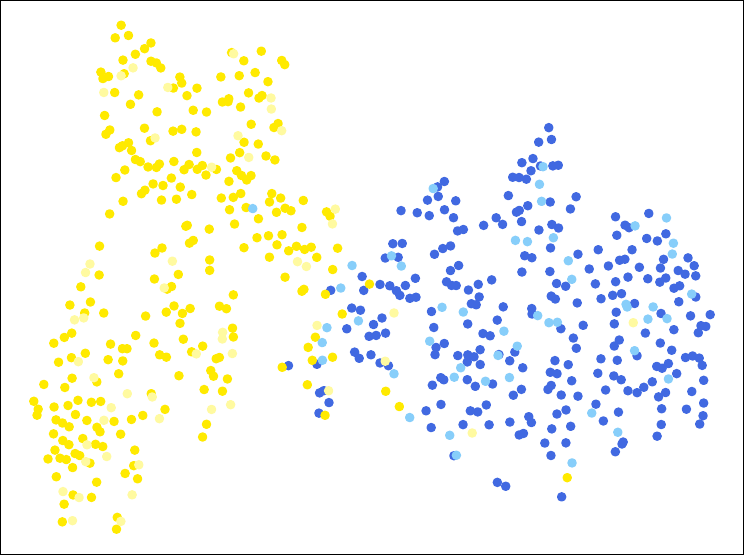}
\\
(c) `Hair' part only \\(most contributing)
\\
\includegraphics[width=\linewidth]{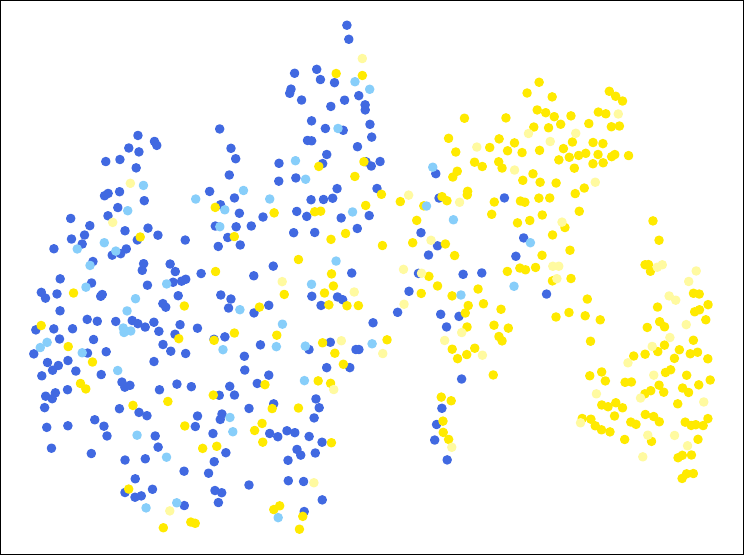}
\\
(d) `Nose' part only \\(least contributing)
\end{minipage}
\begin{minipage}[t]{0.32\linewidth}
\centering
\vspace{0pt}
\includegraphics[width=\linewidth]{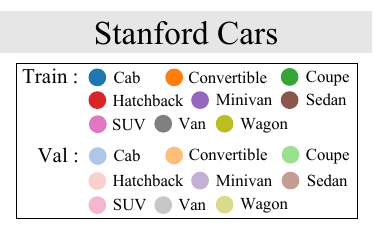}
\\
\includegraphics[width=\linewidth]{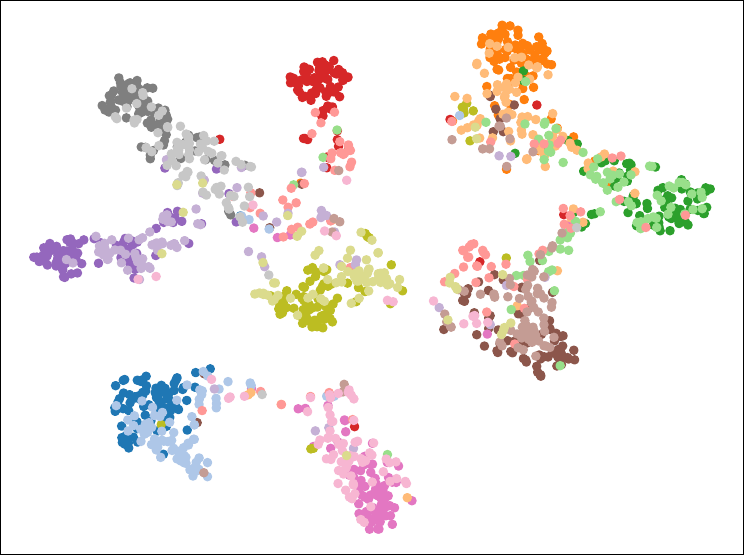}
\\
(e) `Door' part only \\(most contributing)
\\
\includegraphics[width=\linewidth]{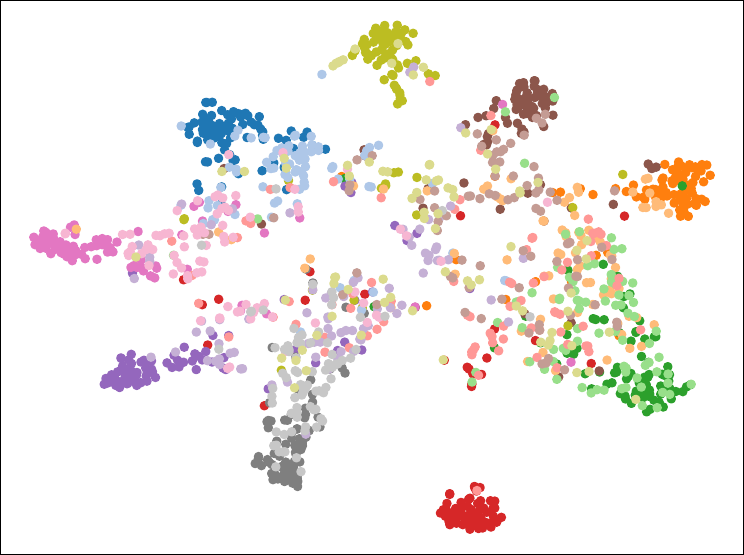}
\\
(f) `Sideglass' part only \\(least contributing)
\end{minipage}

\caption{
\textbf{T-SNE visualization of part features.} 
We visualize the feature spaces of the most contributing part image and the least contributing image using the T-SNE on three datasets: the ASD Screening~\cite{jongmin2022autism} dataset (first row), the SCAT-Drawing dataset (second row), and the Stanford Cars~\cite{krause2013car} dataset (third row). 
%
}
\label{fig:tsne}
\end{figure}

Here, we take a closer look at the feature spaces of the most contributing part images and the least contributing part images on three datasets.
Utilizing T-SNE~\cite{maaten2008tsne}, we examine the feature vectors for each part in isolation on the ASD Screening~\cite{jongmin2022autism} dataset in the first row, on the SCAT-Drawing dataset in the second row, and on the Stanford Cars~\cite{krause2013car} dataset in the third row in \figref{tsne}.
We use the task-level \method{} to select the most and least contributing parts.
The visualization shows a clear pattern: the feature spaces of the most contributing part images, presented in \figref{tsne} (a), (c), and (e), tend to form compact clusters of classes.
In contrast, the feature spaces of the least contributing part images, shown in \figref{tsne} (b), (d), and (f), tend to have more intermingled samples from different classes.
The results demonstrate that the \method{} can effectively identify parts that are crucial for the model decision, as well as the parts that are not crucial for the decision, offering an insightful part-based explanation of a model behavior.

\subsubsection{Robustness to Quality of Part Annotations}
\label{sec:detector-anns}

\begin{figure}[t]
\centering
\footnotesize

\mpage{0.485}{
\includegraphics[width=\linewidth]{figure/img_files/results/class-level/scat/class-level-scat-drawing.pdf}
}
\hfill
\mpage{0.485}{
\includegraphics[width=\linewidth]{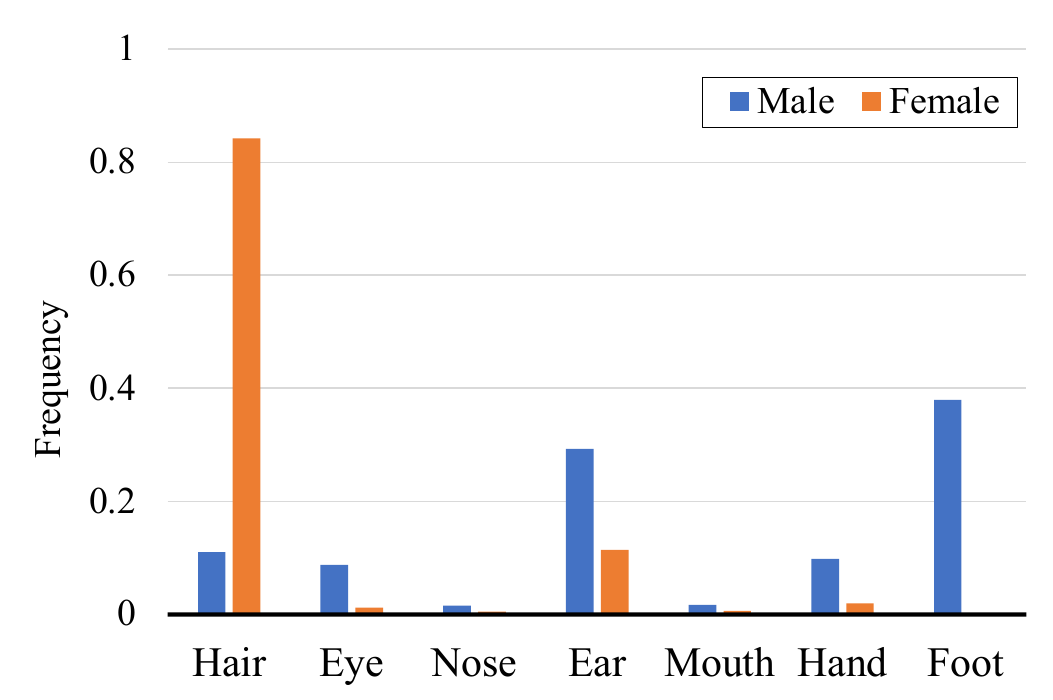}
}
\\
\vspace{-2mm}
\mpage{0.485}{
(a) with G.T. part annotations
}
\mpage{0.485}{
(b) with part detection results
}


\caption{\textbf{
Comparison of part contribution histograms using ground-truth and detector-derived annotations.} 
We show the class-level part contribution histograms generated by the \method{} on the SCAT-Drawing dataset.
(a) The histogram from the \method{} using ground-truth part annotations.
(b) The histogram from the \method{} using part detection results of YOLOv8.
%
}
\label{fig:detector-anns}
\end{figure}

In \figref{detector-anns}, we study the robustness of the \method{} to the quality of part annotations. 
On the SCAT-Drawing dataset, we compare two sets of class-level part contribution histograms:
(a) the male and female class part contribution histograms generated by the \method{} using ground-truth part annotations, 
(b) the male and female class part contribution histograms generated by the \method{} using predicted annotations from an off-the-shelf part detector, YOLOv8\footnote{We use the YOLOv8 implementation provided in the following repository: \url{https://github.com/ultralytics/ultralytics}}.
The histograms from (a) and (b) are quite similar with an average cosine similarity of $0.99$.
The results indicate that the \method{} does not require expensive ground-truth part annotations when a high-quality part detector is available.

\section{Conclusions}
\label{sec:conclusions}

In this paper, we propose the \method{}, a novel framework for explaining models for human figure drawing (HFD) assessment. 
The \method{} explains a model decision by evaluating the contributions of individual parts of an input image based on the Shapley Value. 
It offers more straightforward part-based explanations than previous pixel-level attribution-based methods, \ie a part contribution histogram. 
Our \method{} framework can also generate class/task-level explanations of a model by aggregating the sample-level results.
With the aggregated part-based statistics, we can obtain a more comprehensive understanding of model behavior. 
Moreover, we move beyond the HFD assessment tasks and apply the \method{} on a photo-realistic fine-grained visual classification task.
We rigorously validate the proposed method via extensive and carefully designed experiments on multiple datasets.
%

Our approach relies on part annotations, whether derived from ground-truth annotations or off-the-shelf detectors. 
Concerns may arise regarding the annotation cost and quality. 
As a future work, we plan to devise a training methodology enabling the model to automatically discover parts in an unsupervised manner, thereby integrating the part discovery into our evaluation pipeline.
Also, leveraging our part-based statistics in conjunction with language models enables the generation of textual descriptions. 
The combined utilization of visual histograms and textual descriptions could provide users with more detailed and plausible explanations, thereby enhancing the practicality of our approach.
\section*{Data availability}
No new data were created or analysed during this study. Data sharing is not applicable to this article. The code and pretrained models will be made publicly available upon acceptance.

\section*{Acknowledgement}
This work was supported in part by the Institute of Information and Communications Technology Planning and Evaluation (IITP) grant funded by the Korea Government (MSIT) (Artificial Intelligence Innovation Hub) under Grant 2021-0-02068, (No.RS-2022-00155911, Artificial Intelligence Convergence Innovation Human Resources Development(Kyung Hee University)), and Electronics and Telecommunications Research Institute(ETRI) grant funded by ICT R\&D program of MSIT/IITP[2019-0-00330, Development of AI Technology for Early Screening of Infant/Child Autism Spectrum Disorders based on Cognition of the Psychological Behavior and Response].
This work used datasets from `The Open AI Dataset Project (AI-Hub, S. Korea)'. 
All data information can be accessed through `AI-Hub (\url{www.aihub.or.kr})'.
{\small
\bibliographystyle{plainnat}
\bibliography{main}
}

\end{document}